\documentclass[lettersize,journal]{IEEEtran}
\usepackage{color}
\usepackage{booktabs}
\usepackage[ruled,linesnumbered]{algorithm2e}
\usepackage{graphicx}
\graphicspath{{./images/}}
\usepackage{hyperref}
\usepackage{multirow}
\usepackage{makecell}
\usepackage{cite}
\usepackage{amsmath}
\usepackage{flushend}
\usepackage{tabularray}
\usepackage{threeparttable}
\usepackage{booktabs}
\usepackage{balance}
\usepackage{soul}
\sethlcolor{yellow}
\soulregister{\cite}7
\soulregister{\citep}7
\soulregister{\citet}7
\soulregister{\ref}7
\soulregister{\pageref}7

\usepackage{caption}
\begin{document}

\title{TrafficMCTS: A Closed-Loop Traffic Flow Generation Framework with Group-Based\\ Monte Carlo Tree Search}

\author{Ze Fu$^{\ast}$, Licheng Wen$^{\ast}$, Pinlong Cai$^{\dagger}$, Daocheng Fu, Song Mao, and Botian Shi

\thanks{Ze Fu is with Shanghai Artificial Intelligence Laboratory (this work was done during his internship) and KU Leuven, Belgium. (Email: ze.fu@kuleuven.be)}
\thanks{Licheng Wen, Pinlong Cai, Daocheng Fu, Song Mao, and Botian Shi are with Shanghai Artificial Intelligence Laboratory, Shanghai, China. (Email: wenlicheng@pjlab.org.cn, caipinlong@pjlab.org.cn, fudaocheng@pjlab.org.cn, maosong@pjlab.org.cn, shibotian@pjlab.org.cn)}

\thanks{$^{\ast}$ Equal contribution. $\dagger$ Corresponding author.}
}

\maketitle

\begin{abstract}
Traffic flow simulation within the domain of intelligent transportation systems is garnering significant attention, and generating realistic, diverse, and human-like traffic patterns presents critical challenges that must be addressed. Current approaches often hinge on predefined driver models, objective optimization, or reliance on pre-recorded driving datasets, imposing limitations on their scalability, versatility, and adaptability. In this paper, we introduce TrafficMCTS, an innovative framework that harnesses the synergy of group-based Monte Carlo tree search (MCTS) and Social Value Orientation (SVO) to engender a multifaceted traffic flow with varying driving styles and cooperative tendencies. Anchored by a closed-loop architecture, our framework enables vehicles to dynamically adapt to their environment in real time, and ensure feasible collision-free trajectories. Through comprehensive comparisons with state-of-the-art methods, we illuminate the advantages of our approach in terms of computational efficiency, planning success rate, intention completion time, and diversity metrics. Besides, we simulate multiple scenarios to illustrate the effectiveness of the proposed framework and highlight its ability to induce diverse social behaviors within the traffic flow. Finally, we validate the scalability of TrafficMCTS by demonstrating its capability to efficiently simulate diverse traffic scenarios involving numerous interacting vehicles within a complex road network, capturing the intricate dynamics of human-like driving behaviors.
\end{abstract}


\section{Introduction}

The rapid advancement of intelligent transportation systems (ITS) and autonomous driving technology has significantly transformed urban mobility, increasing the need for large-scale testing and validation to ensure reliability in diverse and complex environments. However, real-world testing is expensive, time-consuming, and often impractical due to safety constraints and limited scenario coverage~\cite{coifman2017critical,caesar2020nuscenes,sun2020scalability}. This has driven the demand for high-fidelity traffic simulation, which enables the development and evaluation of ITS in a controlled yet realistic manner~\cite{barcelo2010fundamentals,lopez2018microscopic}.

Delving deeper into the studies of traffic simulations, a key challenge lies in simulating realistic and diverse driving behaviors that accurately capture human-like interactions in multi-agent traffic environments~\cite{zhu2018human,tong2023humanlike}. Many traffic flow generation methods either rely on simplified rule-based models or pre-recorded datasets, which limit adaptability and generalization in dynamic road conditions~\cite{hidas2005modelling,Chen2023Milestones}. Data-driven approaches~\cite{zhong2023guided,zheng2024genad} have gained attention to capture the complex dynamics of traffic flows, but they often embed driving intentions implicitly rather than defining them explicitly, making individual behaviors difficult to interpret and control~\cite{Fang2024Survey}. Moreover, they face the challenge of unstable trajectory generation and limited long-horizon planning capabilities~\cite{tu2025role}.

Existing open-source traffic simulation tools, such as SUMO \cite{lopez2018microscopic} and CARLA \cite{Carla2017}, have shown their effectiveness in generating traffic flows, but they also have notable limitations. SUMO can support users in customizing road networks and implementing large-scale traffic flow simulations, yet it lacks detailed handling of vehicle movements and interactions within traffic scenarios. CARLA provides open digital assets in the scenarios and supports flexible specifications of sensor suites and environmental conditions. However, CARLA’s behavior modeling is relatively simplified, and its efficiency for large-scale traffic simulations is quite low. Moreover, both SUMO and CARLA rely on simplistic driving models with deterministic parameters, which fail to capture the complexity and variability of human driving behaviors. As a result, the simulated traffic flows tend to display homogeneous driving behaviors, lacking the necessary nuance. Furthermore, these tools are not adequately refined to handle vehicle lateral dynamics \cite{zhang2022trajgen}.

To solve this problem, multi-vehicle trajectory planning has much attention in these years, which often optimizes specific objective functions, such as minimizing travel time or maximizing safety for multiple vehicles \cite{Lenz2016,fisac2019hierarchical, Fan2018b}. 
However, the existing multi-vehicle trajectory planning methods struggle to handle a large number of vehicles in more complex traffic scenarios, which limits these methods to scale up to large-scale traffic flow generation because of overloaded computing resources and time, making them impractical for real-time traffic simulation.
Recent studies, such as InterSim \cite{sun2022intersim}, have relied heavily on pre-recorded trajectory datasets, hampering their potential to generate dynamic traffic flow over long periods due to the limited duration of the trajectories in the datasets.

To address these limitations, we propose TrafficMCTS, a novel traffic flow simulation framework based on group-based Monte Carlo tree search (MCTS). Our framework is designed to generate realistic, diverse, and interpretable traffic flows while maintaining controllability and long-horizon consistency. TrafficMCTS enables both sequential and parallel decision-making among vehicles, efficiently managing large-scale traffic interactions across various traffic scenarios. The main contributions of this paper are as follows:
\begin{itemize}
\item We propose a novel framework called group-based MCTS, which makes decisions in groups formed based on the possibility of interaction between vehicles. This framework can avoid the exponential increase in exploration difficulty in the MCTS. Compared to other methods, our approach significantly reduces computation while maintaining higher planning success rates and more efficient intention completion.

\item We incorporate Social Value Orientation (SVO) to adapt the cooperative tendency of each vehicle to different values, leading to different driving styles within the same scenario. This approach allows us to introduce diversity into the traffic flow at the social interaction level while maintaining interpretability and controllability.

\item We present a closed-loop architecture for traffic flow generation that allows vehicles to react to the dynamic environment in real time while providing collision-free and constraint-satisfying trajectories. 
Unlike SUMO, our approach takes into account vehicle kinematics and offers flexible right-of-way priorities to vehicles, resulting in more realistic traffic flows. Our architecture is capable of handling up to 80 vehicles simultaneously planning on a road network while maintaining an overall simulation acceleration ratio of over 110\%.

\end{itemize}

\section{Related Work}
\label{sec:related work}
This section presents a review of the relevant literature on traffic flow generation. We first introduce some mainstream traffic simulators and then summarize the research work on decision-making and planning for vehicles.

\subsection{Traffic Simulator}
Traffic simulators can provide an effective means of studying traffic phenomena and facilitating the development of autonomous driving technologies. Over the years, some traffic simulators have gained widespread attention and have their own characteristics.

\subsubsection{Simulation for Traffic Management}
Traffic simulations have been developing for decades. Typically, PTV Vissim \cite{fellendorf1994vissim} is a commercial microscopic traffic simulation system for modeling highway and urban traffic. SUMO \cite{lopez2018microscopic} is an open-source traffic simulator that can provide high-performance simulation from a single intersection to an entire city. These simulators for traffic management can improve traffic efficiency by exploring adjustments to traffic strategies, such as optimizing signal timing schemes. However, the depictions of vehicle motion are too simplistic and lack authenticity.

\subsubsection{Simulation for Vehicle Dynamics}
CarSim~\cite{lozoya2012control} provides high-fidelity vehicle dynamics models for virtual testing, simulating braking, handling, ride, stability, and acceleration. 
TNO's PreScan~\cite{tideman2013simulation} integrates hardware-in-the-loop simulation with vehicle dynamics and microscopic traffic simulation with nuanced driver behavior at tactical and strategic levels.
Chen et al.~\cite{chen2019novel} test self-driving algorithms, especially the decision-making process, using a novel hardware-in-the-loop simulation system.
Above simulators primarily target vehicle dynamics tests, assuming human drivers manage challenging tasks. Yet, autonomous vehicles must independently handle these complex driving tasks, including decision-making and planning~\cite{wang2022verification}.
CARLA is an open-source software dedicated to automatic driving research \cite{Carla2017}, with the traffic manager module to generate viable commands for all vehicles in the vehicle registry according to the simulation state. Overall, these simulators can provide realistic simulation environments or vehicle dynamics simulations, but the generation of traffic flow often relies on external tools or model integration.

\subsubsection{Closed-loop Simulation}
In closed-loop simulators, the complex interactions between traffic participants and their environments, coupled with diverse maneuvers, have become the primary focus of research~\cite{huang2022survey}.
Recently, data-driven and learning-based approaches have demonstrated great potential for imitating real-world driving behaviors and reactions through natural driving data. 
TrafficSim \cite{suo2021trafficsim} captures a variety of driver behaviors from human demonstration and parameterizes a joint actor policy that generates socially consistent tracks for all vehicles in the scene. 
SimNet \cite{bergamini2021simnet} presents an end-to-end learning system that responds to the ego vehicle's behavior by predicting the vehicle's motion directly with the bird's eye view. InterSim \cite{sun2022intersim} explains the interaction relationships between the vehicles in the scene and generates trajectories for each vehicle that responds to the test trajectory of the ego vehicle. Flow \cite{wu2017flow} integrates SUMO with the deep reinforcement learning library and enables the development of reliable controllers for complex mixed-autonomy scenarios.
OpenCDA-ROS~\cite{Zheng2023Opencda} merges the OpenCDA framework and ROS to bridge the gap between simulation and real-world deployment, enhancing research, development, and deployment of CDA for cooperative perception, decision-making, and smart infrastructure.
LimSim~\cite{wen2023limsim} is a long-term interactive multi-scenario traffic simulator, which aimed to provide a continuous simulation capability under the complex urban road network. Although closed-loop simulation holds significant importance for evaluating the performance of autonomous driving algorithms, existing simulators still fail to effectively bridge the gap between multi-vehicle dynamic interaction simulation and real traffic conditions, which limits the practical utility of current simulators.

\subsection{Decision-making and Planning}

The decision-making and planning methods for autonomous driving, crucial for generating vehicle flow, have been extensively researched. 
These methods are usually divided into behavioral decision-making, local motion planning, and feedback control. However, the boundaries between these modules are often blurred, with variations appearing in the literature~\cite{Paden2016survey}.

\subsubsection{Decision-making and Planning for Single Vehicles}
A simple but effective approach is applying single-vehicle trajectory planning for each vehicle in the flow. 
Hybrid A* algorithm is a practical path-planning algorithm used in the DARPA Urban Challenge \cite{montemerlo2008junior}. It takes into account the kinematic constraints of the vehicle and captured the continuous vehicle state in the discrete nodes of A* search, which further improves the quality of the solution through non-linear optimization. 
Urmson et al.~\cite{urmson2008autonomous} propose a 3-layer planning system for urban driving including mission, behavioral, and motion planning. Mission planning determines the street-level route, behavioral planning controls lane-changings and intersection right-of-way priorities, while motion planning selects obstacle-avoidance actions. 
Howard and Kelly \cite{Howard2007} propose an approach that efficiently searches the continuum of control space for an optimal trajectory using parameterized controls and nonlinear programming, which is applied to arbitrary terrain shapes and vehicle models actuated in arbitrary ways. 
EM planner~\cite{Fan2018b} introduces a real-time hierarchical motion planning system that covers both highway and lower-speed city scenarios. It employs the Frenét frame for lane-level motion planning and optimizing path and speed simultaneously based on an EM-type iterative algorithm. 
EPSILON \cite{ding2021epsilon} presents a spatiotemporal semantic corridor structure to uniformly define different types of semantic elements such as obstacle, constraint, and path cost, and then solve a general quadratic programming problem to generate a safe and constraint-satisfied trajectory.

\subsubsection{Decision-making and Planning for Multiple Vehicles} In addressing the complexity of multi-vehicle traffic flows composed of heterogeneous agents, researchers have proposed a variety of planning methods. These approaches can broadly be categorized into rule-based methods and learning-based methods.

Rule-based methods can effectively model vehicle intentions and behaviors by understanding the underlying motion mechanisms. Fisac et al.~\cite{fisac2019hierarchical} proposed a hierarchical game-theoretic approach to capture interactions between autonomous vehicles and human drivers. Hang et al.~\cite{hang2022driving} further developed a differential game-based framework for human-like driving at unsignalized intersections, incorporating a Gaussian potential field for collision risk assessment and an event-triggered model for efficiency. Lenz et al.~\cite{Lenz2016} introduced a cooperative motion planning algorithm using the Intelligent Driver Model (IDM) and a cost function for synchronized decisions without inter-vehicle communication. Cai et al.~\cite{cai2022general} used cubic polynomials for trajectory generation, guided by the desired safety margin to mimic safe driving behavior. However, these models are typically designed for simpler scenarios and struggle to scale to dynamic, large-scale interactions.

Learning-based methods, such as imitation learning and multi-agent reinforcement learning (MARL), have become powerful tools for capturing complex driving behaviors, which can leverage large-scale datasets to learn behavior patterns and policy networks that adapt to diverse traffic scenarios. For example, Li et al.~\cite{li2023simulation} introduced a data-driven approach using deep learning and game theory to model background vehicle behavior for autonomous vehicle testing in virtual scenarios, closely simulating real-world interactions. Chen et al.~\cite{Chen2022} proposed a method to train autonomous driving policies using collective fleet experiences and multi-modal sensor data to orchestrate motion planning and ego vehicle trajectory forecasting. The IntentNet framework proposed by Casas et al.~\cite{Casas2021} uses a deep neural network to reason about high-level behaviors and trajectories, incorporating semantic road maps and uncertainty propagation. Sun et al.~\cite{sun2022m2i} presented the M2I framework, which integrates marginal and conditional predictors for multi-agent trajectory forecasting, with a relation predictor to disentangle prediction space. However, these models often suffer from poor generalization, large data requirements, and lack of interpretability, limiting their applicability in safety-critical closed-loop simulations.
 
Recent studies~\cite{tu2025role} have explored generative methods like diffusion-based approaches for realistic simulation. Zhong et al. \cite{zhong2023guided} present CTG, a conditional diffusion model combining controllability and realism for traffic generation. Zheng et al. \cite{zheng2024genad}  propose GenAD, a generative end-to-end autonomous driving framework using a scene tokenizer, variational autoencoder, and temporal model for motion prediction and planning. Liao et al.\cite{liao2024diffusiondrive}  introduce DiffusionDrive, a truncated diffusion policy with multi-mode anchors and a cascade decoder for real-time driving action generation. Yang et al. \cite{yang2025trajectory} leverage Large Language Models (LLMs) to efficiently generate realistic vehicle trajectories from textual descriptions of interactions, thereby enhancing downstream prediction models and improving performance.
Despite these advances, such models still struggle to enforce hard physical constraints (e.g., collision avoidance) and offer limited support for long-horizon planning or interactive decision logic, which is crucial for stable traffic flow generation. Additionally, they face challenges in terms of controllability, as driving intentions are often only implicitly modeled rather than explicitly set, leading to poor interpretability. In long-term scenarios, these models may experience a significant drop in generation quality when encountering rare cases, and they lack the ability to self-correct.

In summary, although existing methods provide diverse solutions for traffic flow simulation, they commonly face three critical limitations: (i) difficulty scaling to large-scale interactions due to computational or sample inefficiencies, (ii) ambiguous representation of driving intent that impairs interpretability and control, and (iii) insufficient support for long-horizon, multi-step interactive decision-making. These limitations hinder their effectiveness in generating realistic, controllable, and consistent traffic flows for simulation-based validation. An ideal solution should balance scalability, interpretability, and physical fidelity, while providing robust support for long-term interaction modeling. Li et al.~\cite{Li2022} present an MCTS-based planning algorithm enriched with predictive heuristics and a social-compliant reward system, facilitating preference identification. However, the existing MCTS-based approach has not fully addressed key challenges such as scalability to dense multi-agent scenarios and realism in human-like interaction modeling.
In this context, our work explores a novel group-based planning paradigm grounded in the MCTS-based approach, aiming to address these challenges.

\begin{figure*}[tbp]
    \centering
    \includegraphics[width=.75\textwidth]{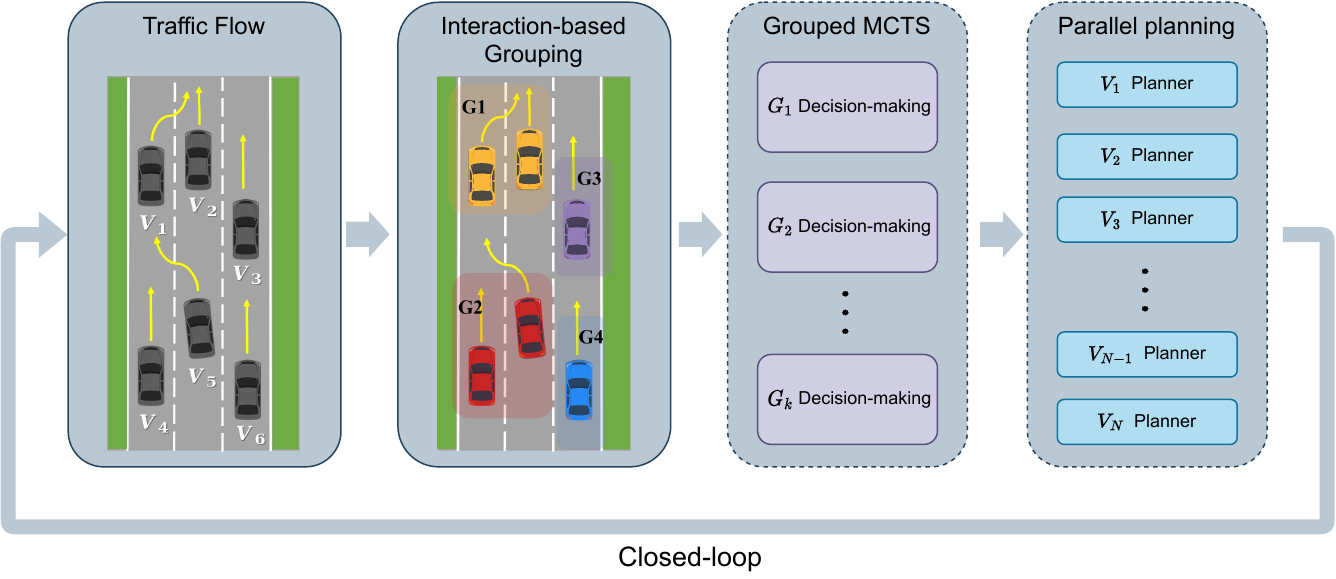}
    \caption{\rmfamily The architecture of TrafficMCTS. The yellow arrows indicate the driving intentions of each vehicle.}
    \label{fig:framework}
    \vspace{-10pt}
\end{figure*}

\section{Overview}
\label{sec:framework}
\subsection{Problem Formulation}

The setup and goal of the traffic flow generation problem are clarified below. 
This work supports multi-vehicle joint decision-making with defined driving intentions. Each vehicle’s routing information, assumed to be available, serves as the basis for its driving intention, guiding it toward its destination while mitigating congestion. The driving intentions can be classified into several types, as illustrated in Table \ref{tab:intentions}. 
The objective is to generate feasible trajectories that follow each vehicle’s driving intention, enabling appropriate lane changes and turns to facilitate smooth traffic flow toward destinations. Vehicles must avoid collisions and exhibit diverse yet reasonable social interactions.
Considering the uncertainty of interaction behavior at intersections, this work does not focus on intersection scenarios.

\begin{table}[tbp]
    \caption{\rmfamily Driving intentions for vehicles.}
    \centering     
    \begin{tabular}{c c}
    \hline
    Driving intention & Description	\\
    \hline
    Keep\_Lane          & Maintain the current lane	\\
    Change\_Lane\_Left  & Drive into the left lane	\\
    Change\_Lane\_Right & Drive into the right lane	\\
    Merge\_In           & Merge in from the ramp 	\\
    Drive\_out          & Drive out into the ramp    \\
    Overtake            & \makecell[c]{Overtake the vehicle ahead and  return the\\ current lane} \\
    \hline
    \end{tabular}
    \label{tab:intentions}
\end{table}

Considering the scenario in which the traffic flow is generated, suppose there are $n$ vehicles to be controlled by our system, denoted as $\mathcal{V} = \{V_1,V_2,\cdots,V_n \}$, where $n>0$.  
Meanwhile, there could be $m$ vehicles driving independently in the scenario, which is manipulated by other algorithms or controlled by humans, $m\geq0$. Considering a vehicle $V_i\in\mathcal{V}$, $\mathbf{x}^i_t$ denotes its state at time $t$, such as position and heading angle. A feasible trajectory of vehicle $V_i$ should:
\begin{itemize}
    \item Be defined as $\pi_i= \left[\mathbf{x}^i_0 ,\mathbf{x}^i_1,\cdots, \mathbf{x}^i_T\right]$ , where $T$ is the total duration of simulation. The final state at the end of simulation $\mathbf{x}^i_T$ must be on the vehicle's destination or on its way to its destination. 
    \item Satisfy the vehicle dynamics $\mathbf{x}_{t+1}^i = f\left(\mathbf{x}_{t}^i,u^i_t\right)$, where $u^i_t$ is the control input at time $t$ of $V_i$, such as throttle and steering angle, and $f$ represents vehicle's dynamic function.
    \item Not collide with each other at any time, $\textit{shape}\left(\mathbf{x}_{t}^i\right) \cap\textit{shape}\left(\mathbf{x}_{t}^j \right) = \emptyset, i\neq j ,\forall t \in \left[0, T \right]$,  where $\textit{shape}()$ denotes the curve function of the vehicle body. 
\end{itemize}

\subsection{System Architecture}

To solve the traffic flow generation problem mentioned above, we propose TrafficMCTS, a closed-loop traffic flow generation framework that considers different behavioral preferences among vehicles. Compared with the open-loop strategy, our framework enables interaction between vehicles, making it more suitable for generating long-term traffic flows without distribution shift. As shown in Fig. \ref{fig:framework}, the architecture includes four key modules and they form a closed loop generation of traffic flow: scenario cognition based on driving intentions, interaction-based grouping, grouped MCTS, and parallel planning.

The TrafficMCTS framework takes the states of the entire traffic flow and the driving intentions of each vehicle as inputs. The core of the framework is first to make a rough but long-term joint decision for vehicles, followed by parallel trajectory planning based on that decision. However, the large number of vehicles in the traffic flow leads to a large search space and low computational efficiency for simultaneous joint decision-making. To address this, we propose a group-based MCTS method that divides traffic flow into several decision groups based on the potential interaction between vehicles. For each decision group, the vehicles make joint decisions without priority, and the decision order between groups is classified into two types: sequential and parallel. 
After this process, each vehicle possesses a decision based on its state, and the TrafficMCTS framework uses a parallel trajectory planner to generate an optimal trajectory that guarantees safety.
Finally, the vehicle positions within the traffic flow are updated according to the planned trajectories. Above steps are repeated to complete the closed-loop process of traffic flow generation.

We first describe the details of the group-based MCTS approach in Section \ref{sec:grouped MCTS}.
Secondly, the closed-loop process of TrafficMCTS is more intricate than described above, which involves distinct closed-loop frequencies for the decision-making process and the planning process. It will be elaborated further in Section \ref{sec:Closed-loop Decision and Planning}.
In addition, the architecture of TrafficMCTS supports the setting of unique driving preferences for vehicles to increase the diversity of traffic flows, which will be further discussed in Section 
\ref{sec:flow diversity}.

The source code of our proposed traffic flow generation framework is available as part of the open-sourced project LimSim \cite{wen2023limsim}. Please refer to \url{https://github.com/PJLab-ADG/LimSim}.

\section{Methodology}
\label{sec:methodology}
\subsection{Group-based Monte Carlo Tree Search}
\label{sec:grouped MCTS}

Multi-vehicle decision-making is always a challenging problem, because as the number of vehicles increases, the computational performance and decision success rate decrease. To address this issue, we propose a novel approach that groups vehicles based on their interaction and makes decisions separately for each group. In this section, we first introduce the multi-vehicle MCTS algorithm, which can handle decision-making problems among multiple vehicles in traffic flow. We then describe our vehicle grouping design based on interaction analysis, which aims to reduce the computational complexity of the decision-making process. Finally, we present our approach to making decisions in groups, which makes the multi-vehicle MCTS algorithm more efficient.

\subsubsection{Multi-vehicle MCTS}

The MCTS method can effectively solve the problems with huge exploration space. Starting from the initial node at $t = 0$, the general MCTS algorithm performs $k$ iterations with four phases: selection, expansion, simulation, and back-propagation. 
The selection phase adopts the upper confidence bounds for trees (UCT) policy \cite{browne2012survey} to address the exploration-exploitation dilemma. 
In the expansion phase, a collision check will be performed to prune the conflict child node. From a valid node, the algorithm uses a default policy to perform the simulation until it reaches a terminal node. In the back-propagated phase, the rewards from the simulation sample will be used to update the traversed nodes’ rewards. After a certain number of iterations, the search tree returns the node with the best reward.
When applying the MCTS method to multi-vehicle joint decision-making, the prevailing methodology follows the approach of setting priorities for vehicles. Vehicles with higher priorities make decisions first, and those with lower priorities obey previous decisions.

\begin{figure}[tbp]
    \centering
    \includegraphics[width= .7\linewidth]{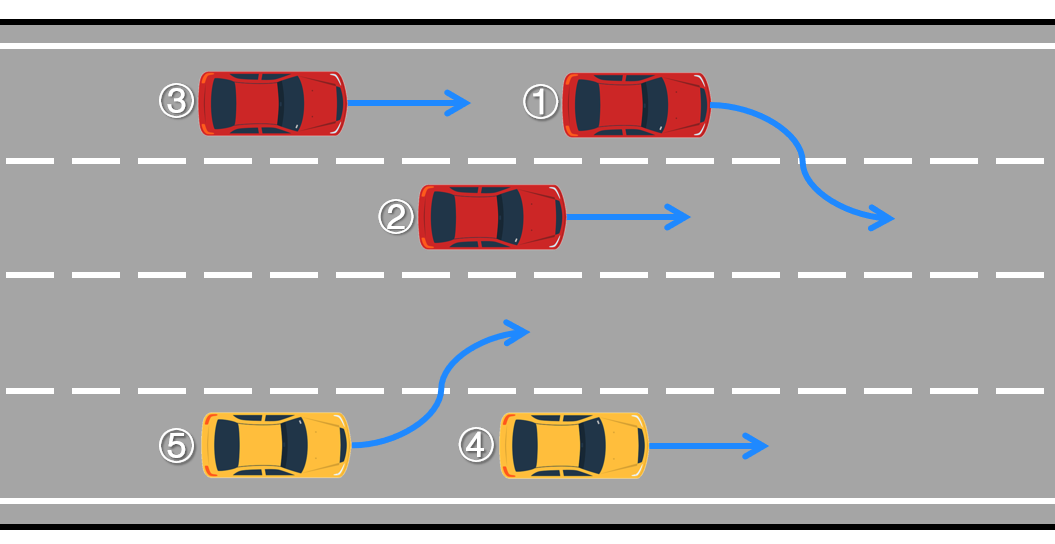}
    \caption{\rmfamily Schematic of the multi-vehicles with driving intentions in the highway scenario.}
    \label{fig:freeway_demo}
    \vspace{-10pt}
\end{figure}

\begin{figure*}[tbp]
    \centering
    \includegraphics[width= .8\textwidth]{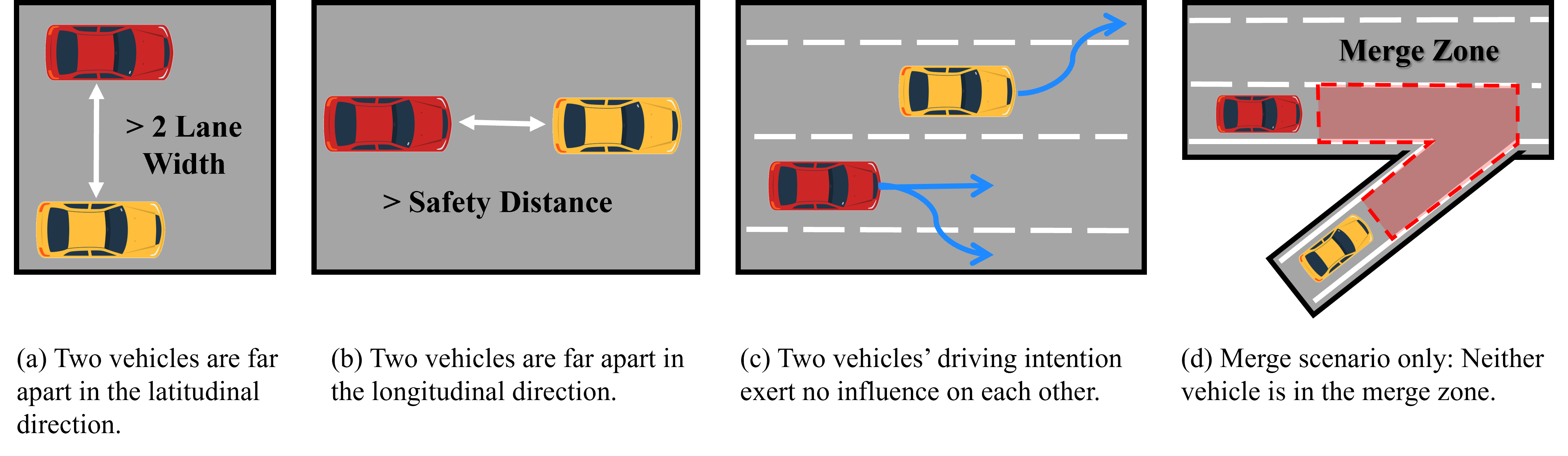}
    \caption{\rmfamily Cases in which vehicles are \textbf{not} considered to interact with each other.}
    \label{fig:no_inter}
\end{figure*}

\begin{table*}[tbp] 
    \centering
    \caption{\rmfamily Actions available for vehicles in one-time step}
    \resizebox{0.75\textwidth}{!}{
        \begin{tabular}{c c c}
        \toprule
            Action & State change                                                            & Description                                   \\ \midrule
            KS     & $\left[v\Delta t,0,0\right]$                                            & \makecell[l]{Maintain the current velocity}   \\
            AC     & $[v\Delta t+a_{acc}\Delta t^2/2,0,a_{acc}\Delta t]$  & \makecell[l]{Increase the current velocity    with a fixed acceleration $a_{acc}$} \\
            DC     & $[v\Delta t+a_{dec}\Delta t^2/2,0,a_{dec}\Delta t]$ & \makecell[l]{Decrease the current velocity    with a fixed deceleration $a_{dec}$} \\
            LCL    & $\left[v\Delta t,-\Delta d,0\right]$                                    & \makecell[l]{Make a partial left lane-changing   with a width of $\Delta d$} \\
            LCR    & $\left[v\Delta t,\Delta d,0\right]$                                     & \makecell[l]{Make a partial right lane-changing  with a width of $\Delta d$}  \\
            \bottomrule
        \end{tabular}
    }
    \label{tab:actions}
\end{table*}

In our previous work \cite{Wen2023}, we introduced the concept of \textit{metanode} to replace the standard node in the general MCTS. This metanode enables the generation of simultaneous actions for multiple vehicles, thereby eliminating the need for priority assignment during the decision-making process. As shown in Table \ref{tab:actions}, there are five available actions for vehicles at each time step. At time step $t$, all controlled vehicles within the metanode select one of the five actions, while uncontrolled vehicles in the flow are assumed to perform a lane-keeping (KL) action. These chosen actions induce changes in the state of the traffic flow, leading to the formation of a new metanode. A terminal metanode indicates either all controlled vehicles within it have completed their intentions or that no feasible solution could be found in the decision period.

In this paper, we incorporate information about the lane $l_{idx}$ in which the vehicle is currently traveling into the vehicle's state to accommodate different scenarios. Thus, the state for vehicle $V_i$ at time step $t$ is described as $\mathbf{x}_i^t = [s, d, v, l_{idx}]^T$. During the expansion phase, the lane-change check is performed on the metanode by combining the vehicle's Frenét coordinates with road information, such as lane width and ramp length. This ensures that the values of $d$ and $l_{idx}$ are dynamically updated to maintain their correct relationship.

Using such a search method, all controlled vehicles can perform their intended driving actions based on the provided routing details.
However, as the number of vehicles in the scenario increases, the search tree expands exponentially due to the joint consideration of all vehicles' possible behaviors. Consequently, this expansion leads to a significant reduction in computational efficiency and overall success rate.
To enhance the performance of the multi-vehicle MCTS algorithm for large-scale traffic flows in complex scenarios, we propose a novel group-based multi-vehicle decision-making method as below.

\subsubsection{Vehicle Grouping based on Interaction Analysis}

It is neither practical nor efficient to require each vehicle to be aware of all other vehicles, especially for vehicles that are too far apart to interfere with one another. 
For instance, considering the traffic flow depicted in Fig. \ref{fig:freeway_demo}. Vehicle 1, with the intention to change lanes to the right, will primarily interact with its neighboring vehicles 2 and 3 within a certain decision horizon. However, the available actions of vehicle 1, such as maintaining the current lane (AC) or executing a lane-changing to the right (LCR), have little effect on vehicles 4 and 5. 
Even in scenarios where human-driven vehicles (HVs) and autonomous vehicles (AVs) coexist, human drivers are limited by their visual perception and attention span, and they typically focus on the surrounding vehicles. This observation inspires us to divide the global interaction problem into a set of sub-problems. We partition the traffic flow into local groups based on the possibility of interaction, allowing for separate decision-making within each group. 
This approach enables us to maintain the benefits of the multi-vehicle MCTS algorithm, including high interpretability and interactivity, while significantly reducing the size of the solution space.

Several grouping methods have been developed to address the complexity of the joint multi-vehicle decision-making problem. Desiraju et al. divide the problem of maximizing safe lane changing on a highway with multiple lanes into subproblems on a three-lane highway \cite{Desiraju2015CAVs}. Within each subproblem, a grouping of vehicles is conducted based on the separation distances between vehicles, which are evaluated according to vehicles' positions and speeds. Li et al. propose a grouping method to meet the mandatory lane-changing demands for connected and automated vehicles \cite{Li2020CAVs}. This method primarily considers the distances between vehicles and imposes a maximum limit on the number of vehicles within each group. Although these approaches considered interactions between vehicles, they are limited to highway scenarios and could only make lane-changing decisions. Due to the simple grouping strategies, they did not fully exploit the potential benefits of grouping, leading to the over-filtering of feasible solutions.

In this paper, we propose a grouping algorithm based on interaction analysis, aiming to group vehicles that are likely to interact with each other into the same group whenever possible. This approach takes into account the vehicles' intentions mentioned in Table \ref{tab:intentions}, allowing for a more refined judgment of interaction between vehicles. The grouping method based on the interaction judgment goes beyond the highway scenario and offers a more interpretable solution. By utilizing the driving intentions derived from the routing information, the interaction between vehicles can be detected more accurately. We discuss the possibility of interaction between two vehicles $V_i$ and $V_j$ in the next several time steps. In some cases, they may not interact with each other, because they are far from each other or their actions taken based on their driving intentions will exert little influence on each other. These cases are summarized and listed in Fig. \ref{fig:no_inter}. Especially, the Merge Zone is defined as the section of the ramp and its nearest lane on the main road, extending a distance of $l$ ahead of the merge point. To avoid the potential collision in the next $k$ time steps, vehicles $V_i$ and $V_j$ should maintain the safety distance $s_d$, which is defined by 

\begin{equation}
    \label{eq:safety_distance}
    \begin{aligned}
            s_d\! =\!\begin{cases}
                \!&\!(v_j - v_i)\!\cdot\!kt + \frac{1}{2}(a_{\text{acc}}-a_{\text{dec}})\!\cdot\!(kt)^2 + MSD, \\
                &\!\quad\text{if } v_j \geq v_i \\
                \!&\!MSD,\!\quad\text{if } v_j < v_i
            \end{cases}
    \end{aligned}
\end{equation}

where $MSD>0$ is constant, representing the minimum safe distance.
$s_d$ represents the distance from the front of the rear vehicle to the rear of the front vehicle, and is calculated based on the assumption that the leading vehicle $V_i$ consistently performs a DC (Decelerate and Change lane) action with $a_{\text{dec}}$, while the following vehicle $V_j$ consistently performs an AC (Accelerate and Continue in the same lane) action with $a_{\text{dec}}$ within the next $k$ time steps.

We use an adjacency matrix to record the information about the interaction possibilities between vehicles, which is called the interaction matrix. 
For a given traffic flow $\mathcal{V}$ consisting of $n$ vehicles, the interaction matrix $\Phi$ has a size of $n \times n$. Especially, if $\Phi_{i,j} = \Phi_{j,i} = 1$, it indicates that vehicles $V_i$ and $V_j$ are likely to interact with each other. Conversely, if $\Phi_{i,j} = \Phi_{j,i} = 0$, it suggests that there is no expected interaction between them. Since exchanging subscripts does not affect the determination of conflicting relationships, we only consider $\Phi_{i,j}$ in the following text and assume that $\Phi_{j,i}$ can be automatically updated with $\Phi_{i,j}$. 

\begin{algorithm}[tbp]
\small
\SetAlgoLined
 \textbf{Initialize}~The interaction matrix $\Phi$ with all elements as 0; \\
 \tcp{Sort vehicles by their longitudinal coordinates decreasingly}
 $\mathcal{V}_s \gets \mathcal{V}$ \\
 \For{$V_i\leftarrow V_1$ \KwTo $V_n$}{
    \For{$V_j\leftarrow V_{i+1}$ \KwTo $V_n$}{
        \tcp{Judge the relationship by the driving intentions}
        \eIf{$\left(V_i, V_j\right){\rm \ in\ No \ Interaction \ Cases}$ }{
            $\Phi_{i,j} \gets 0$;
        }{
            $\Phi_{i,j} \gets 1$;
        }
        \tcp{Overtaking intention requires additional consideration}
        \If{$\left(V_i, V_j\right){\rm \ is\ an\ Overtake\ Pair}$ }{
            $\Phi_{i,j} \gets 1$;
        }
    }
 }
 \KwResult{The interaction matrix $\Phi$.} 
 \caption{Judge\_Interaction($\mathcal{V},\Psi$)}
 \label{algo:judge_inter}
\end{algorithm}

Algorithm \ref{algo:judge_inter} outlines the procedure for constructing the interaction matrix $\Phi$, which takes the vehicle flow $\mathcal{V}$ and the driving intentions $\Psi$ of the vehicles as input. 
The algorithm begins by sorting the flow $\mathcal{V}$ in non-increasing order based on the vehicles' longitudinal coordinates. This ensures that for any vehicles $V_i$ and $V_j$ in $\mathcal{V}_s$, if $i < j$, then $s_i \geq s_j$. It is important to note that since different lanes have different Frenét coordinate systems, all vehicles are converted to the same Frenét coordinate system by default in this algorithm.
Next, starting from $V_i$, the algorithm determines whether $V_i$ and $V_j$ fall into any of the cases listed in Fig. \ref{fig:no_inter}. If a match is found, indicating no interaction between the vehicles, the corresponding element in the interaction matrix is set to 0. Otherwise, $\Phi_{i,j}$ are set to 1, indicating a potential interaction between the vehicles.
It is important to highlight that an overtake pair, consisting of one vehicle with the intention to overtake and the other being overtaken, must be marked as having a possibility of interaction. This ensures that they can be grouped together later to facilitate the decision-making process for the Overtake intention.

\begin{algorithm}[tbp]
\small
\SetAlgoLined
 $\textbf{Initialize}~V_1.group\_idx \gets 1$, $G_1 \gets \left\{V_1\right\}$, and $\mathcal{G} \gets \left\{G_1\right\}$;\\
 \For{$V_i\leftarrow V_2$ \KwTo $V_n$}{
    \tcp{Initialize the index of the group for $V_i$ as 0}    
    ${V}_i.group\_idx \gets 0$;\\
    \tcp{Reverse traversal to ensure adjacently grouping}
    \For{$V_j\leftarrow V_{i-1}$ \KwTo $V_1$}{
        \tcp{Not be grouped together if the group is full}
        \If{${\rm len}\left({G}_{V_j.group\_idx}\right) \geq N_{limit}$}{
            continue;
        }
        \tcp{Not be grouped together if no interaction}
        \eIf{$\Phi_{i,j}$ {\rm is} $0$}{
            continue;
        }{
        \tcp{Put $V_i$ into the group to which $V_j$ belongs}
            $V_i.group\_idx \gets V_j.group\_idx$; \\
            $G_{V_i.group\_idx} \gets G_{V_i.group\_idx} \cup \left\{V_i\right\}$; \\
            break;
        }
    }
    \tcp{Form a new group}
    \If{$V_i.group\_idx$ {\rm is 0}}{
        $V_i.group\_idx \gets {\rm len}(\mathcal{G}) + 1$;\\
        $G_{V_i.group\_idx} \gets \left\{V_i\right\}$;\\
        $\mathcal{G} \gets \mathcal{G} \cup \left\{G_{V_i.group\_idx}\right\}$;\\
    }
 }
\KwResult{The grouping result $\mathcal{G}$ composed of several groups.}
 \caption{Grouping($\mathcal{V}_s, \Phi$)}
 \label{algo:grouping}
\end{algorithm}

With the aid of the interaction matrix, the grouping of vehicles can be achieved. 
Algorithm \ref{algo:grouping} outlines the steps involved in the vehicle grouping process.
The algorithm takes the sorted vehicle flow $\mathcal{V}_s$ and the interaction matrix $\Phi$ as input. The maximum number of vehicles allowed in a single group is limited to $N_{limit}$ to control the size of the search tree constructed for each group. The selection of the parameter $N_{limit}$ will be discussed later in Section \ref{sec: make_decision_group}.

Algorithm \ref{algo:grouping} begins by initializing a group containing the first vehicle $V_1$ in $\mathcal{V}$. For each subsequent vehicle $V_i$, the algorithm iterates through the vehicles $V_j$ that are located in front of $V_i$, with $j$ ranging from $i-1$ to 1. Traversing from the back to the front ensures that $V_i$ is grouped with its closest interacting vehicle.
By examining the interaction matrix $\Phi$, if there is a possibility of interaction between $V_i$ and $V_j$, and the number of vehicles in the group to which $V_j$ belongs does not exceed $N_{limit}$, $V_i$ will be grouped with $V_j$. After the traversal, if $V_i$ has not been grouped yet, a new group will be created for it. The output of Algorithm \ref{algo:grouping} is a set of groups $\mathcal{G}$, where each group contains no more than $N_{limit}$ vehicles.
It is worth noting that due to the limit of the maximum number of vehicles in a group, Algorithm \ref{algo:grouping} does not guarantee that vehicles with interaction possibilities are all grouped together. For two groups $G_1 $and $G_2$ in $ \mathcal{G} $, if there is no possibility of interaction between vehicles in $G_1 $ and vehicles in $G_2$, the decision-making process can be conducted in parallel on them. Otherwise, the decision-making process should be sequential for them.

\begin{figure}[tbp]
    \centering
    \includegraphics[width=0.8\linewidth]{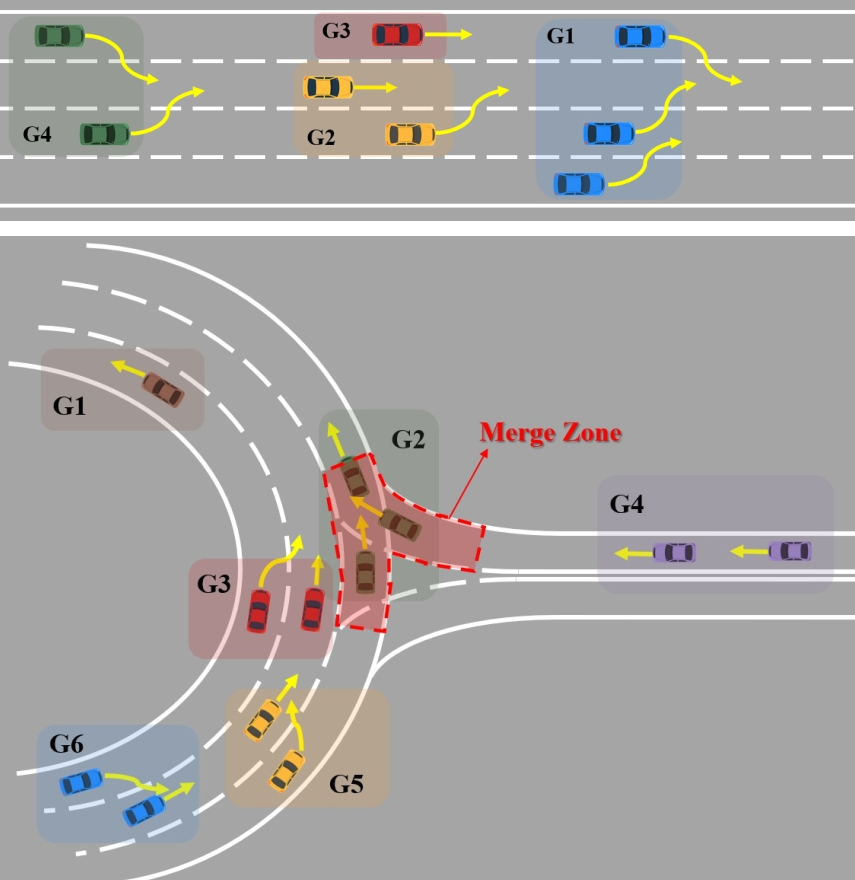}
    \caption{\rmfamily Results of grouping in freeway/roundabout scenario.}
    \label{fig:grouping_demo}
    \vspace{-10pt}
\end{figure}

Fig. \ref{fig:grouping_demo} illustrates the results of grouping in scenarios such as freeways and roundabouts. Our grouping algorithm takes into account detailed vehicle driving intentions, thus grouping vehicles that may interact with each other into the same group as much as possible. Moreover, vehicles without the possibility of interaction will be divided into different groups, which can improve computing efficiency through parallel decision-making and planning.

\subsubsection{Making decision in Groups}

\label{sec: make_decision_group}

Once the grouping process is completed, the decision-making process is conducted. 
Parallel decision-making is permissible if there is no vehicle interaction between two adjacent groups. Conversely, the decision-making process of the following group should take the decision result of the preceding group into consideration, where sub-problems are created and the multi-vehicle MCTS method can be used separately for each group. 

This decision-making order is reasonable since after Algorithm \ref{algo:grouping}, the group with a smaller group number is located in front of the road segment, and the preceding group has the initiative of passage and might have an impact on the following groups. 

\begin{figure}[tbp]
    \centering
    \includegraphics[width = 1.0\linewidth]{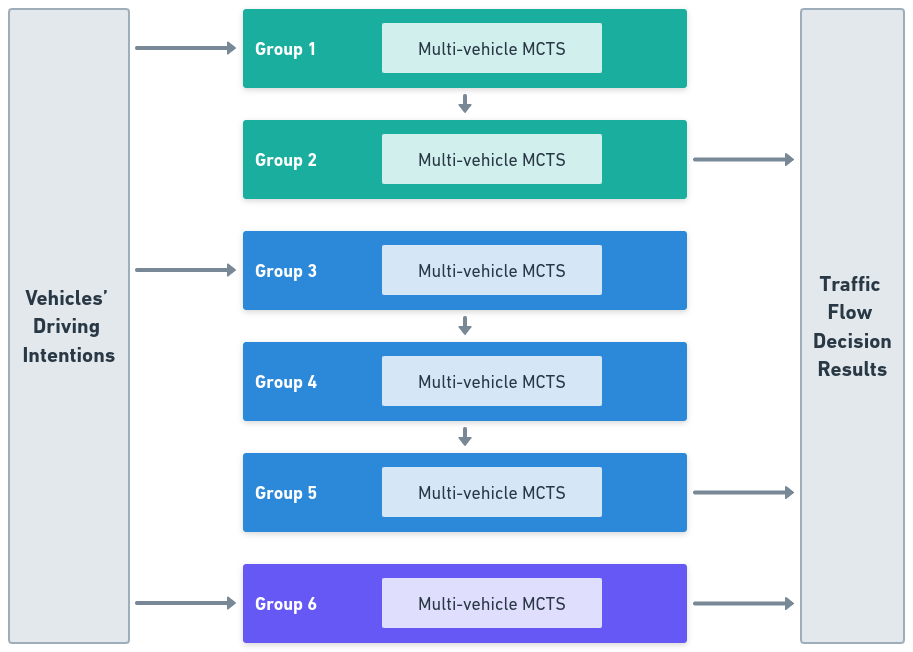}
    \caption{\rmfamily An example of making the decision for a traffic flow with 6 groups of vehicles.  Groups 1 and 2, as well as groups 3, 4, and 5, have inter-dependencies and require sequential decision-making. Group 6 has no inter-dependencies and can be directly decided upon. Therefore, group 6 can perform decision-making simultaneously alongside group 1 and 3.}
    \label{fig:solving_steps}
    \vspace{-10pt}
\end{figure}

Fig. \ref{fig:solving_steps} displays a traffic flow segmented into six distinct groups. 
Due to the interdependence of the vehicles in group 2 on the decisions made by group 1, a sequential approach to decision-making is essential. Similarly, sequential decision-making is required for vehicles in groups 3, 4, and 5. In contrast, vehicles in group 6 make decisions independent of the decisions of any other group, allowing them to engage in parallel decision-making alongside the above groups.

For the $k$-th group in the sequential decision-making process, the multi-vehicle MCTS method takes the flow $\mathcal{V}_k$ as input, encompassing vehicles within the current group $G_k$ as well as those from all preceding groups $G_1, G_2, \dots, G_{k-1}$. The time taken by vehicle $V_i$ in the earlier flow to finalize its decision is denoted as $t_i$. As the tree expands up to time step $t$, actions and states of $V_i$ from the previous group's decision results are directly retrieved if $t \leq t_i$. Conversely, if $t$ surpasses $t_i$, actions and states of $V_i$ are acquired from the prediction module, which is a simple IDM in this study. The states of all HVs in the metanode are also obtained via the prediction module. Once the search tree concludes the decision-making for $G_k$, the outcomes and time steps for accomplishing driving intentions are documented. Following this, vehicles within $G_k$ are integrated into the previous flow. This iterative process continues until decisions are made for all groups in sequential decision-making process.

\begin{figure}[tb]
    \centering
    \includegraphics[width=1.0\linewidth]{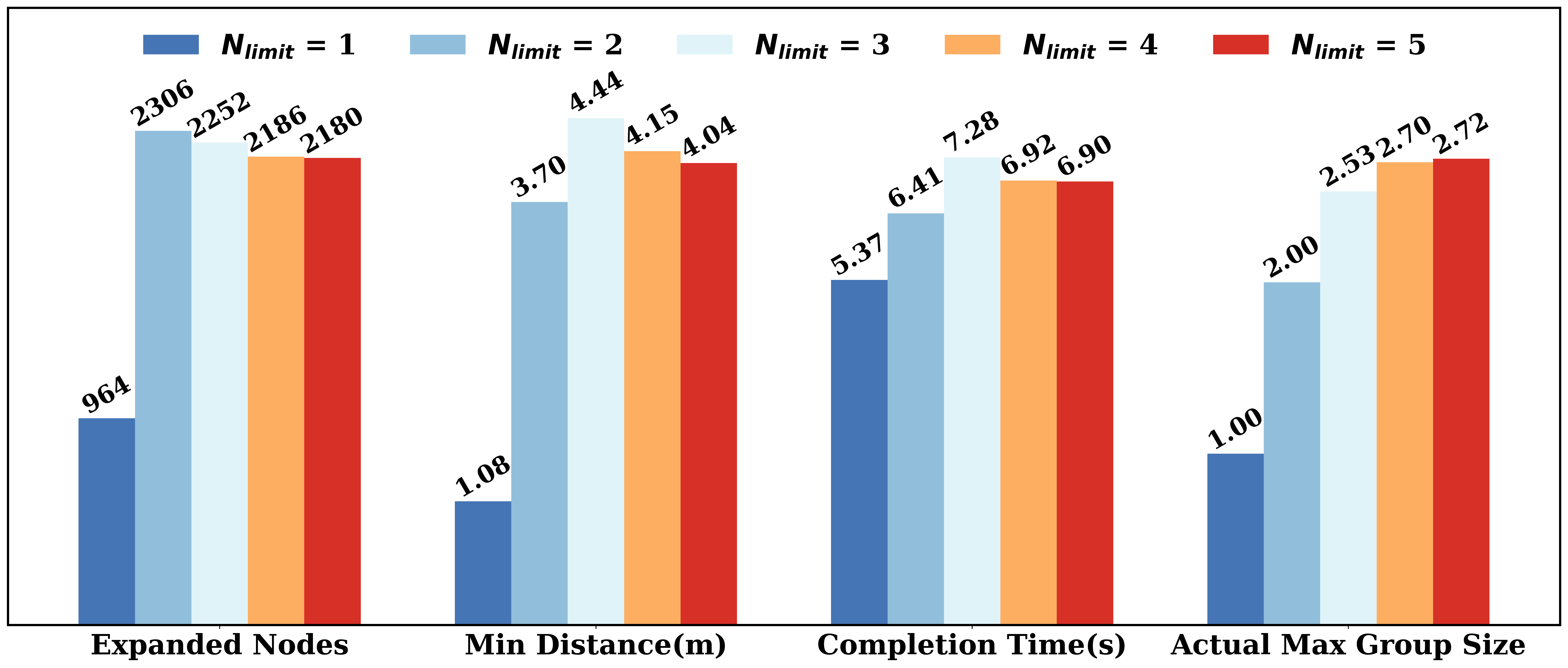}
    \caption{\rmfamily \textcolor{black}{Experimental indicators of the test at different maximum group size limitations. 
    The indicators in the horizontal axis (from left to right) are: (1) average expanded nodes in MCTS, (2) average minimum distance between vehicles, (3) average time taken for all AVs to complete their driving intentions, (4) average actual maximum size of the groups.}}
    \label{fig:N_limit_res}
    \vspace{-10pt}
\end{figure}

Before utilizing the aforementioned approach for decision-making, it is crucial to determine the appropriate maximum vehicle number $N_{limit}$ within each group. We test the group-based MCTS algorithm in a freeway scenario containing a four-lane expressway to explore the effect of the selection of $N_{limit}$ on the algorithm performance.
In this scenario, there are a total of 10 vehicles distributed across a 70-meter section, with at least six of them having the intention to change lanes. The decision module operates with a time step of 1.5 $s$, and the assessment of interaction possibilities considers the subsequent two time steps ($k=2$ in Eq. (\ref{eq:safety_distance})).
The maximum of time steps is set as 10, which means that the search tree can expand up to the state at a maximum of 15 $s$. To guarantee the scalability of our approach and the effectiveness of the grouping strategy, $N_{limit}$ ranges from 1 to 5, and each test is repeated ten times under same experimental setup. 
Results are shown in Fig. \ref{fig:N_limit_res}.

As $N_{limit}$ increases, the average number of expended nodes of the search tree increases firstly, then gradually decreases and tends to stabilize. Similarly, the average time taken for each group to complete their driving intentions follows the same trend. This can be attributed to the fact that when the number of vehicles in a group is small, especially when $N_{limit}=1$, the search tree terminates quickly. However, during the search process, interactions between vehicles are ignored, resulting in a very small average minimum distance between vehicles in the decision results. This means that the vehicles adopt extremely dangerous driving actions when completing their driving intentions. 
Additionally, it is observed that as $N_{limit}$ varies from 1 to 5, the average maximum number of vehicles in one group stabilizes below 3. This indicates that under the computational efficiency constraints, setting $N_{limit}$ to 3 can fulfill the need for grouping potential interacting vehicles together as much as possible.
At the same time, indicators such as the average expansion nodes and the average minimum distance perform well at $N_{limit}=3$. Therefore, $N_{limit}$ is set to be 3 in our implementation of the group-based MCTS.

\subsection{Closed-loop Decision-making and Planning}
\label{sec:Closed-loop Decision and Planning}

In this section, we focus on the closed-loop decision-making and planning process in our proposed framework. This process involves two key components: re-decision and replanning. These two components enable our framework to adapt to dynamic traffic scenarios and make real-time decisions and plans for safe and efficient traffic generation.

\subsubsection{Re-decision}
\label{sec:re-decision and re-grouping}

In Section \ref{sec:grouped MCTS}, we propose a group-based MCTS method to generate long-term coarse decisions for large-scale traffic flows with complex interactions. The method consists of two main steps: grouping and multi-vehicle MCTS. 
The grouping algorithm focuses on the interactions between vehicles in certain time steps. However, for a flow dynamically changing, the interactions between vehicles also change constantly, so the grouping results need to be updated regularly. 
Furthermore, in some scenarios where the interactions between vehicles are complex, the group-based MCTS may not provide a feasible solution for all vehicles to complete their driving intentions within the decision interval. Therefore, re-decision is particularly necessary for the decision module to respond to the dynamic environment.

The interval of re-decision, denoted as $T_{update}$, plays a crucial role in the effectiveness of the decision-making process. If $T_{update}$ is too small, frequent updates of groupings and decisions may disrupt the vehicle’s execution of former decisions, leading to discontinuous driving behavior. This can lead to a reduced success rate for the flow in completing their driving intentions. 
Conversely, if $T_{update}$ is too large, the grouping results may no longer match the interactions between vehicles after several time steps. In this situation, the expansion node of the search tree cannot focus on handling the interaction between adjacent vehicles, which also reduces the search efficiency and decision success rate. A better choice is to make $T_{update}$ vary with the success rate of the decision results, instead of making it a fixed time interval. The decision success rate $\gamma$ refers to the ratio of the number of controlled vehicles completing their driving intention $N_{finish\_decision}$ to the number of controlled vehicles $N_{need\_decision}$. Mathematically, we define that
\begin{equation}
    \gamma = N_{finish\_decision} / N_{need\_decision}.
\end{equation}
Then, we define the re-decision period $T_{update}$ as
\begin{equation}
    T_{update}=T_{min} + \gamma\cdot \left(T_{max}-T_{min} \right),
    \label{eq:T}
\end{equation}
where $T_{min}$ and $T_{max}$ represent the lower and upper limits of  $T_{update}$, respectively. It indicates that when $\gamma$ is high, our framework tends to ensure that the vehicles can coherently execute the decision results. Conversely, when $\gamma$ is low, the loop can run frequently to find feasible solutions as soon as possible in the dynamic environment.

In our group-based MCTS method, we check whether each controlled vehicle completes its driving intention. For a controlled vehicle $V_{i}$, if its driving intention is to change lanes, merge, or leave the main road, then the intention can be achieved when it is located in the target lane. If the intention is to overtake, it must also be ahead of the target vehicle $V_{aim}$ to complete the intention.
For vehicles successfully completing their driving intentions in a decision loop, the action sequences will be sent to the planning module to obtain the final feasible trajectories. Otherwise, they will be provided with lane-keeping trajectories by the planning module and wait for the next decision period to obtain feasible solutions.

\subsubsection{Replanning}
The planning module with a parallel architecture is proposed by \cite{Wen2023} to generate continuous, kinematic-feasible, and collision-free trajectories. For each vehicle $V_i $ under our control, the single-vehicle trajectory planner receives an action sequence from the decision module. After predicting the future trajectories of other vehicles in the flow, which is achieved by directly utilizing the planning results from the previous planning period in this work, it splits the different actions in the sequence and a sub-trajectory planner connects all trajectories for each action segment to obtain the final feasible and continuous vehicle trajectory.

The sub-trajectory planner expands the predefined constant state change described in Table \ref{tab:actions} to form a continuous region of target states. In particular, for vehicles executing lane-keeping actions, particularly controlled vehicles that are unable to complete their driving intentions during the decision-making process, the planner generates a larger target state region to facilitate rational in-lane behaviors. 
Subsequently, a set of target states is sampled within this predefined region.

We adopt quintic polynomial curves in the Frenét frame to generate the jerk-optimal connections between the start state and all sampled target states of the sub-trajectory. These alternative sub-trajectories are calculated with a cost function, which is detailed in Section \ref{sec:cost set}. The sub-trajectory with the lowest cost that meets the kinematic constraints and is collision-free with other vehicles is chosen. The final state of this sub-trajectory is then established as the initial state for the next sub-trajectory.

The above process is repeated in the fixed time interval $T_{planning}$ to avoid a potential collision. In each loop, the current state of $V_i$ will be set as the initial state for the next trajectory. In addition, the driving intention completion judgment mentioned in Section \ref{sec:re-decision and re-grouping} is conducted in every re-planning loop based on the vehicles’ current states. In this way, the driving intentions of the flow are updated timely when entering the next decision-making loop.

\subsection{Generation of Diverse flow}
\label{sec:flow diversity}

One of the key objectives of our multi-vehicle decision-making and planning framework is to foster traffic flows with increased diversity. This is achieved by integrating the SVO \cite{Schwarting2019} into the decision module and introducing the trajectory weight set within the planning module. By considering various cooperative tendencies and driving habits among vehicles and appropriately setting the SVO values and trajectory weight sets, we can effectively capture and reflect their individual characteristics. In this way, while attaching diversity to the flow, we can also control the behaviors of vehicles.

\subsubsection{Decision Reward with SVO Embedded}
During the simulation phase of group-based MCTS, the algorithm continues to select the valid metanode to extend the next time step until the simulation terminates. Then a reward will be calculated for this simulation from the result. We designed a SVO embedded reward function that guides vehicles to efficiently complete their driving intentions while characterizing their cooperative tendencies. 
Inspired by \cite{Schwarting2019}, the reward $R_i$ for $V_i$ in the traffic flow consists of reward to self $R_{i,self}$ and reward to other vehicles $R_{i,others}$, which are weighted by SVO angular preference $\varphi_i$. The definition of reward $R_i$ is shown as
\begin{equation}
{\color{black} 
    R_i = \cos \left(\varphi_i \right) \cdot R_{i,self} + \sin \left(\varphi_i \right) \cdot R_{i,others},
}    
    \label{eq:SVO}
\end{equation}
where the orientation $\varphi_i \in [0, 2\pi]~rad$. In the traffic flow, each vehicle would possess its unique factor $\varphi_i$. Some common definitions of the social preferences are
\begin{itemize}
    \item Altruistic ($\varphi_i \approx \pi /2$): Maximize the others' rewards, without consideration of their own outcomes.
    \item Prosocial ($\varphi_i \approx \pi /4$): Behave to benefit the whole group and maximize the joint reward.
    \item Individualistic/egoistic ($\varphi_i \approx 0$): Maximize their own outcomes, without concern for others' rewards.
\end{itemize} 

The reward to self $R_{i,self}$ focuses on the efficient and safe completion of $V_i$’s driving intention, which includes two parts: the reward for the terminal state $R_{i,terminal}$ and the reward for the process states $R_{i,process}$. 
\begin{equation}
{\color{black} 
    R_{i,self} = R_{i,terminal} + R_{i,process},
}
    \label{eq:reward}
\end{equation}
The former only concentrates on the leaf metanodes of the search tree, while the latter focuses on the path from the initial metanode to the terminal metanode.
$R_{i,terminal}$ is exclusively acquired when $V_i$ completes its driving intention in the terminal metanode. $R_{i,process}$ is obtained by evaluating the state of the flow in each time step from multiple perspectives and adding them up. To improve travel efficiency, $V_i$ receives higher rewards when its speed approaches its target speed. To ensure rational driving behavior, $V_i$ receives higher rewards for continuous actions and staying on the center line of the lane. Furthermore, to enhance safety, $V_i$ receives higher rewards for maintaining a larger distance from its surrounding vehicles. 

The reward $R_{i,others}$ is concerned with the interaction between $V_i$ and other vehicles in the flow. It is a negative value, i.e., it is the penalty caused by improper interactions. 
Improper interactions encompass actions such as recklessly lane-changing behavior, on-ramp vehicles rushing to merge into the main road, and main road vehicles refusing to yield to on-ramp vehicles. For $V_i$ with the intention of Change\_Lane\_Left, Change\_Lane\_Right or Overtake, it receives a penalty when its lane-changing actions force the rear vehicle in the target lane to decelerate. That is, $V_i$ chooses LCL or LCR action and the rear vehicle chooses DC action simultaneously in a metanode. 
Vehicles in the Merge Zone will be penalized for improper behavior, such as on-ramp vehicles trying to merge while main road vehicles refuse to give way.

After calculating $R_{i,self}$ and $R_{i,other}$, the reward $R_i$ for $V_i$ can be obtained by Eq. (\ref{eq:SVO}). It’s worth noting that the reward $R_i$ should meet in the range of 0 to 1.
As the search tree in group-based MCTS handles all vehicles in the same group, the reward of the simulation is obtained by combining all rewards of vehicles within the terminal metanode as
\begin{equation}
    R_{group} = \frac{1}{K}\underset{1\leq i \leq K}{\sum}R_i,
    \label{eq:total_reward}
\end{equation}
where $K$ is the number of vehicles in the group and $ R_{group}$ is the average reward of vehicles.
Since $R_i \in [0, 1]$, the distribution of $R_{group}$ is also guaranteed to stay in $[0, 1]$. Finally, $R_{group}$ is used to update the average rewards of all traversed metanodes in the back-propagation phase.

\subsubsection{Cost Function for Planning}
\label{sec:cost set}

The planner we employ bears resemblance to the one suggested by \cite{werling2010optimal}. It employs uniform sampling of target states determined by the chosen action and generates jerk-optimal trajectories by utilizing quintic polynomial curves connecting the initial state to all the sampled target states.
To assess the safety and comfort of each alternative trajectory, it is essential to calculate a corresponding cost function.

In particular, there are several cost terms for the vehicle $V_i$ in the function:
\begin{itemize}
  \item Curve energy cost: $J_{i,cur} = \sum_{t \in \Gamma} \left(\kappa_i(t)\right)^2$, $\kappa_i(t)$ denotes the curvature of a point in the trajectory $\Gamma$. It penalizes the trajectory with sharp turns.
  \item Heading difference cost: $J_{i,\phi} = \sum_{t \in \Gamma} \left|\phi_i (t)  - \phi_r \right|^2$, $\phi_i(t)$ denotes the orientation angle of the vehicle at a point in the trajectory $\Gamma$ and $\phi_r$ indicates the road's direction at the point.
  \item Out of line cost: $J_{i,out} = \sum_{t \in \Gamma} \left| d_i(t) \right|^2$, $d_i(t)$ denotes the lateral offset in Frenét frame at a point in the trajectory $\Gamma$. This term is only active for actions except for lane-changing ones.
  \item Acceleration cost: $J_{i,acc} = \sum_{t \in \Gamma} \left(a_i(t)\right)^2$, $a_i(t)$ denotes vehicle's acceleration at a point in the trajectory $\Gamma$.
  \item Jerk cost: $J_{i,jerk} = \sum_{t \in \Gamma} \left(j_i(t)\right)^2$, $j_i(t)$ denotes vehicle's jerk (a derivative of acceleration) at a point in the trajectory $\Gamma$. This term and the acceleration cost both contribute to the comfort of the trajectory.
  \item Dynamic obstacle cost: $J_{i,obs}$. This term serves to prevent the vehicle from being too close to the surrounding vehicles, which is described in detail below.
\end{itemize}

The dynamic obstacle cost takes into account the distance between a vehicle following a trajectory and dynamic obstacles on the road, such as surrounding vehicles. The cost for vehicle $V_i$ is defined as
\begin{eqnarray}
J_{i,obs} &=& \sum_{t \in \Gamma} \sum_{\forall j\neq i}J_{i,obs}(t,j),
\end{eqnarray}
where $J_{i,obs}(t,j)$ represents the cost related to the position of vehicle $V_j$ relative to the current vehicle $V_i$ at time step $t$.
In this study, we classify the position of vehicle $V_j$ relative to $V_i$ into three categories: 1) outside the alert zone; 2) inside the alert zone without collisions; 3) experiencing collisions with the current position of $V_i$. When vehicle $V_j$ is located outside the alert zone of $V_i$, it is considered safe and incurs no additional cost. Conversely, direct collisions with the trajectory of $V_i$ are deemed unacceptable and will be assigned an infinite cost. Within the alert zone, the cost increases as vehicle $V_j$ approaches closer to $V_i$. For more details, please refer to our previous work \cite{Wen2023}.

Aforementioned terms are combined into a cost function:
\begin{eqnarray}
J_i &=& \sum~w_{(\cdot)}~J_{i,(\cdot)},
\end{eqnarray}
where $J_{i,(\cdot)} \in \left\{J_{i,cur},~J_{i,\phi},~J_{i,out},~J_{i,acc},~J_{i,jerk},~J_{i,obs}\right\}$ represents the various cost components mentioned earlier, and $w_{(\cdot)}$ denotes the weight parameter corresponding to each cost component. It is worth noting that setting $w_{obs}$ to a small value can increase collision risk to generate corner cases.

We observe that different drivers in a traffic flow may select different trajectories based on their driving preferences. In other words, drivers assign varying weights to these trajectory cost components. For example, some drivers may prioritize driving comfort, while others may prefer maintaining a greater safe distance from other vehicles.
To account for this, we define a set of trajectory weights, denoted as $\mathcal{W}$, which includes $M$ weight vectors based on driver habits:
\begin{eqnarray}
\mathcal{W} = \left\{\mathbf{w}_1,\mathbf{w}_2,\cdots,\mathbf{w}_M \right\}.
\end{eqnarray}
Each weight vector $\mathbf{w}_k \in \mathcal{W}$ contains a unique combination of weights for all the cost components, namely $\mathbf{w}_k = \left[w_{cur},~w_{\phi},~w_{out},~w_{acc},~w_{jerk},~w_{obs}\right]$.
Each vehicle within the traffic flow individually selects a unique weight vector from the set $\mathcal{W}$ to generate its customized trajectories.

\section{Experiments}
\label{sec:experiment}
The simulation experiments are conducted to assess the performance of our proposed framework. First, a case study is presented to show the improvement of our group-based MCTS algorithm to other relevant approaches. Then, we illustrate how our closed-loop framework can bring diversity to AVs and manage the scenarios involving a mix of AVs and HVs. Finally, long-term simulations are conducted to demonstrate our framework’s ability to generate realistic large-scale traffic flows with complex interactions. In particular, a comparative experiment with SUMO displays the framework's efficient decision-making ability in ramp scenarios.

In the first two experiments, the initial states of the vehicle flow are derived from real-world driving records obtained from the CitySim dataset \cite{ou2022}. Driving intentions such as lane-changings, entering or exiting ramps, and overtaking leading vehicles may be assigned to the vehicles. These driving intentions serve as inputs to the framework. The proposed framework is implemented under the environment of Python 3.10 on a PC with an Intel Core i7-11700 @ 2.50GHz CPU with 8 cores. To facilitate the discussion, the parameters of the vehicles and the reward settings in all experiments are listed in Table \ref{tab:veh_parameters} and Table \ref{tab:reward_settings}, respectively.
The time step is set to 1.5 $s$ in the decision module and 0.1 $s$ in the planning module. The grouping algorithm deals with interaction possibilities within the next two decision time steps. 
Besides, we set the minimum safe distance $MSD = 2~m$ in Eq. (\ref{eq:safety_distance}).
The maximum decision horizon is 9 $s$. In the closed-loop framework, the re-planning period is set to 0.3 $s$, and the lower and upper limits of the re-decision period are set to 1.5 $s$ and 6 $s$, respectively.
For the sake of convenience, the weight of the cost function for vehicle planning remains unchanged across all experiments.
The detailed experiment video can be found on YouTube via {\href{https://youtu.be/Tr0QWCasO54}{https://youtu.be/Tr0QWCasO54}}.

\begin{table}[tbp]
    \caption{\rmfamily Parameters of the vehicles during the decision-making process.}
    \label{tab:veh_parameters}
    \centering 
        \begin{tabular}{ccc}
        \hline
        Symbol & Definition & Value	\\
        \hline
        $w$          & width of the vehicle	                &   $2~m$\\
        $l$          & length of the vehicle                &	$5~m$\\
        $a_{acc}$    & acceleration for available actions   &	$1.0~m/s^2$\\
        $a_{dec}$    & deceleration for available actions   & 	$-1.0~m/s^2$\\
        $\varphi$    & SVO angular preference               &   $\pi/4~rad~\rm{(by~default)}$\\
        \hline
        \end{tabular}
\end{table}

\begin{table}[tbp]
    \centering
    \caption{Reward settings in the tests.}
    \label{tab:reward_settings}
    {
    \begin{tblr}{
      width = 0.98\linewidth,
      colspec = {Q[2.8]Q[8.4]Q[6.6]},
      row{1} = {c},
      cell{2}{1} = {c},
      cell{2}{3} = {c},
      cell{3}{1} = {r=4}{c},
      cell{3}{3} = {c},
      cell{4}{3} = {c},
      cell{5}{3} = {c},
      cell{6}{3} = {c},
      cell{7}{1} = {c},
      cell{7}{3} = {c},
      hline{1,8} = {-}{0.08em},
      hline{2} = {-}{0.05em},
      hline{3,7} = {-}{},
    }
    Type  & Description   & Value      \\
    $R_{i,terminal}$  & Completing driving intention  & 0.8           \\
    $R_{i,process}$ \tnote{1} 
    & Driving on the center line of the lane in $k_l$ steps                                                  & $0.1 {k_l}/{k_t}$                  \\
    & Keeping consistent actions with previous time steps in $k_c$ steps                                         & $0.2 {k_c}/{k_t}$                  \\
    & Maintaining safe distance with fronts vehicle in $k_d$ steps& $0.2 {k_d}/{k_t}$                  \\
    & Approaching the target speed $v_{t}$                     & $0.2\sum\limits_{i=0}^{k_t}\min(v_i/v_t, 1)/k_t$ \\
    $R_{i,others}$ & Acting recklessly for interactions in $k_t$ steps & $-0.1k_r$          
    \end{tblr}
    \begin{tablenotes}
     \item[1] * $k_t$ is total decision steps.  
    \end{tablenotes}
    }
\end{table}

\begin{figure}[tbp]
    \centering    \includegraphics[width=.8\columnwidth]{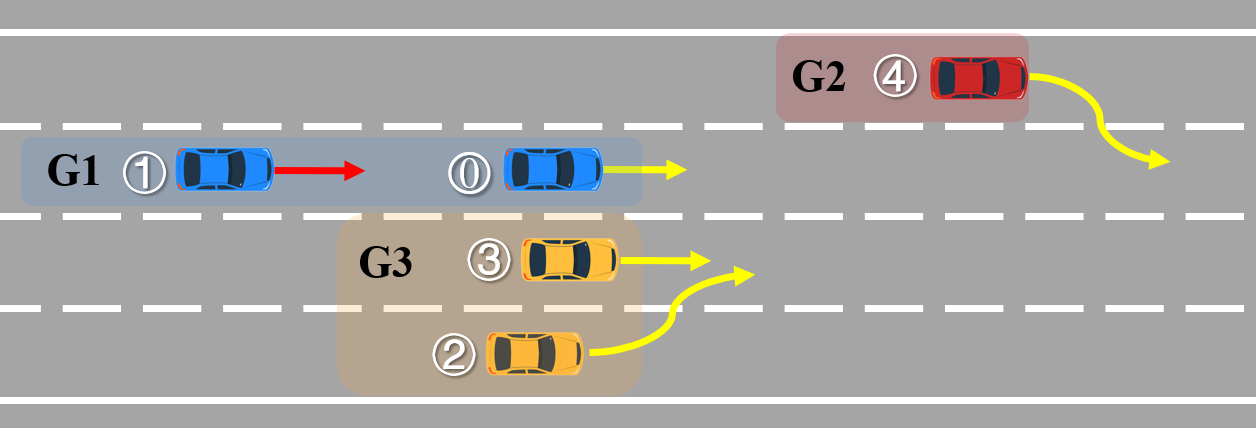}
    \caption{ \rmfamily The four-lane freeway scenario of the test.}
    \label{fig:test2.1_scenario}
    \vspace{-10pt}
\end{figure}

\begin{figure}[tb]
    \centering
    \includegraphics[width = 0.9\linewidth]{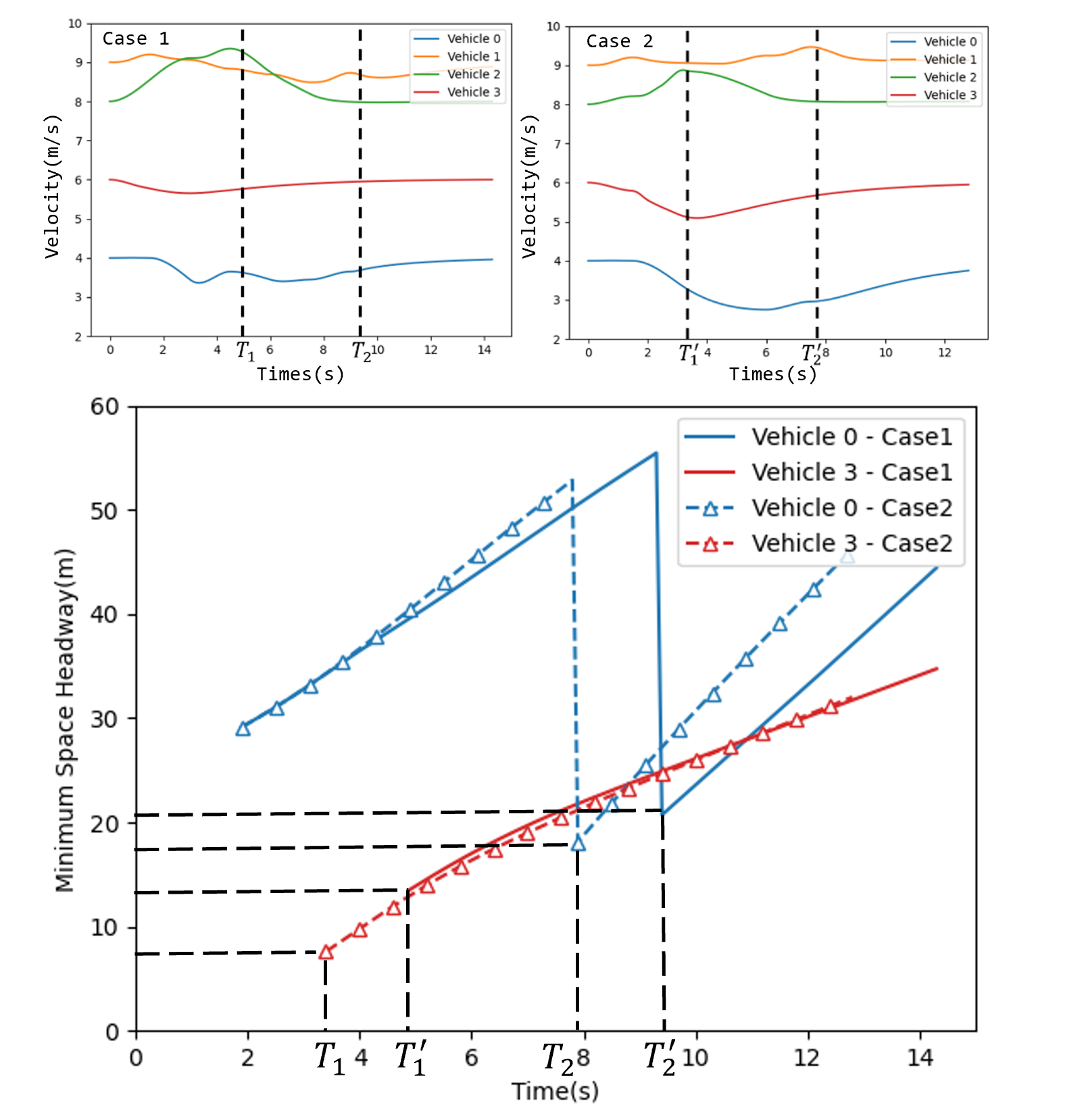}
    \caption{\rmfamily Distances to leading vehicles and the velocity profiles of certain vehicles in the test.}
    \label{fig:test2.1_res}
    \vspace{-10pt}
\end{figure}

\subsection{Comparison of Method Performance}
\label{sec:test1}
In this subsection, the performance of the group-based MCTS method is tested in a ramp scenario, which includes a three-lane main road with an on-ramp merging into it. All the vehicles in the traffic flow are controllable AVs. The number of AVs is set to 3, 6, and 9. To make the situation more complex, vehicles on the side lane of the main road will change lanes towards the middle lane, and vehicles on the middle lane are randomly assigned the driving intentions of Change\_Lane\_Right or Change\_Lane\_Left. The driving intentions of the on-ramp vehicles are Merge\_In. 

We use the sequential MCTS approach \cite{Li2022} as a comparison to evaluate the effectiveness of our proposed method. This method makes decisions for vehicles in order of priority within each decision time step. In this experiment, vehicles ahead possess a higher decision priority. The available sampled acceleration in sequential MCTS is chosen from $\{-1.0, 0, 1.0\}~m/s^2$, and the available lateral sampled velocity is chosen from $\{-1.2, 0, 1.2\}~m/s$.
Moreover, to verify the effectiveness of the grouping algorithm, we also implement an MCTS approach based on the random grouping strategy. 
This approach does not take into account the possibility of interaction between vehicles during the grouping process. Instead, it assigns each vehicle a group index by randomly selecting a value between 1 and the total number of vehicles in the traffic flow $n$.
It ensures that each group possesses at least one vehicle.

The time step and the maximum decision horizon are set the same for all three methods. Under different vehicle numbers, the maximum iteration number of MCTS is set to 2000, 4000, and 6000, respectively. For the group-based MCTS and the random-grouped MCTS, the sum of MCTS iterations for each group should not exceed the maximum iteration number. Each method is repeated ten times under the same experimental setup. The results of the decision-making method will be sent to the same planning module with a planning horizon of 12 $s$ for all three methods.
After that, the success rate of each method, the average number of expansion nodes in MCTS, the average number of available actions for each node in the search tree, the average time to finish the vehicles’ driving intentions, and the minimum inter-vehicle distance in the generated solutions are measured. 
The results are shown in Table \ref{tab:test1}.

\begin{table*}[tb]
    \caption{\rmfamily Comparative experiments with 3/6/9 vehicles in the ramp scenarios.}
    \label{tab:test1}
    \centering
    \resizebox{\textwidth}{!}{%
    \begin{tabular}{l c c c c c}
    \toprule
    Method & Avg. Expanded Nodes & \makecell[c]{Avg. Available Actions\\ for Each Node} & Success rate (\%) & Min. Distance $(m)$  & Avg. Finish Time $(s)$ 	\\
    \midrule
    \makecell[l]{Sequential MCTS}          & $4924.7 / 9228.5 / 13033.2$ & \makecell[c]{$167.3 / 91501.7 /12674568.9$ }& $96.7 / 56.7 / 31.1$  & $17.04 / \textbf{5.07} / \textbf{3.62}$ & $2.88 / 8.80 / 11.73$\\
    \makecell[l]{Random-grouped MCTS}  & $2712.1 / 4137.1  / 5592.9$ & $28.6 / 35.4 / 42.0$           & $100.0 / 80.0 / 81.1 $ & $20.38 / 4.25 / 1.52$ & $2.46 / 5.69/ 5.68$\\
    TrafficMCTS                     & $\textbf{867.9} / \textbf{3001.2} / \textbf{3529.8}$   &	$\textbf{7.9} / \textbf{24.5} / \textbf{28.7}$         & $\textbf{100.0} / \textbf{95.0} / \textbf{88.9}$ & $\textbf{20.39} / 4.25 / 2.45$ & $\textbf{2.32}/ \textbf{3.57}/ \textbf{4.64}$\\
    \bottomrule
    \end{tabular}
    }
\end{table*}

All three methods achieved high success rates in the ramp scenarios with three controlled AVs. However, when the number of AVs reaches 6, the success rate of the sequential MCTS method plummets to $56.7\%$, and the random-group-based MCTS can only attain a success rate of $80\%$, while our method maintains a success rate of $95\%$. When $n$ reaches 9, the vehicle flow in the scenario is relatively congested and the interaction relationship is complex. Our method still manages to achieve a success rate of $88.9\%$, which is significantly better than the success rate of $31.1\%$ of the sequential MCTS method at this time. 
That is because the priority defined by sequential MCTS prevents higher-priority vehicles from yielding to the lower-priority vehicles, thereby filtering out some feasible solutions. On the other hand, as the number of vehicles increases, the search tree without grouping strategy grows exponentially, making it difficult to find a solution within a limited number of iterations.

The search efficiency can be reflected in the average number of expansion nodes. Under three sets of experimental parameters, our method is always the fastest to find a feasible solution among the three methods. Especially when $n=9$, under the same search iteration number limit, the sequential MCTS method expands nearly four times nodes than our method, but the success rate of its solution is far inferior to ours. By adopting the grouping strategy, the average number of available actions for each node remains small with the increase in the number of vehicles. However, this indicator grows exponentially for the method without a grouping strategy. When $n = 9$, this indicator of the sequential MCTS exceeds ten million, reaching an intolerable order of magnitude and leading to a huge decrease in search efficiency and success rate. Besides, compared to the random-group-based MCTS, the expansion nodes of our method focus on handling the interactions between adjacent vehicles. Therefore, even with a similar number of available actions for each node, our method can find a feasible solution more quickly.

In terms of the quality of the solutions, the average time for vehicles to complete their driving intentions is always the shortest in the solutions generated by our method. Meanwhile, the minimum inter-vehicle distances of our method are always good and hence ensure the safety of the vehicle flow. For $n=6$ and $n=9$, this value of our method is slightly smaller than that of the sequential MCTS. It is because in congested cases, the sequential MCTS often fails in decision-making and the vehicle defaults to keep driving in their current lane. But our method can still guide the vehicles safely to complete their lane-changing intentions. It is worth noting that the planning module refines the trajectory further, bringing all these three methods to collision-free trajectories.

\begin{figure}[tb]
    \centering
    \includegraphics[width=.9\linewidth]{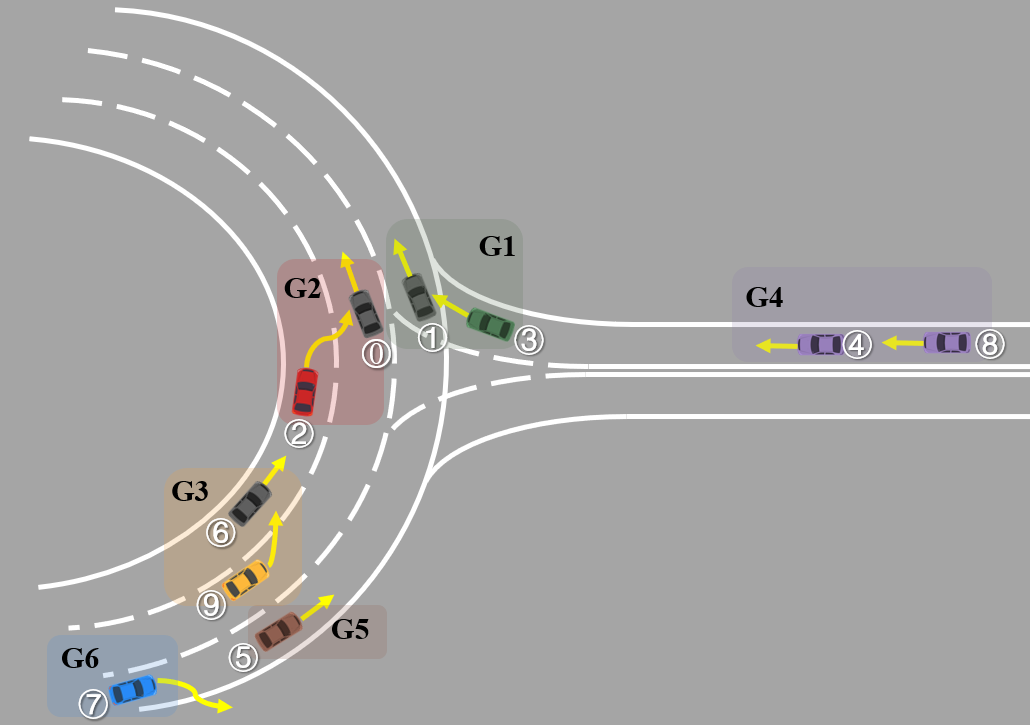}
    \caption{\rmfamily The roundabout scenario of the test.}
    \label{fig:test2.2_scenario}
    \vspace{-5pt}
\end{figure}

\begin{figure}[t]
    \centering
    \includegraphics[width=0.98\linewidth]{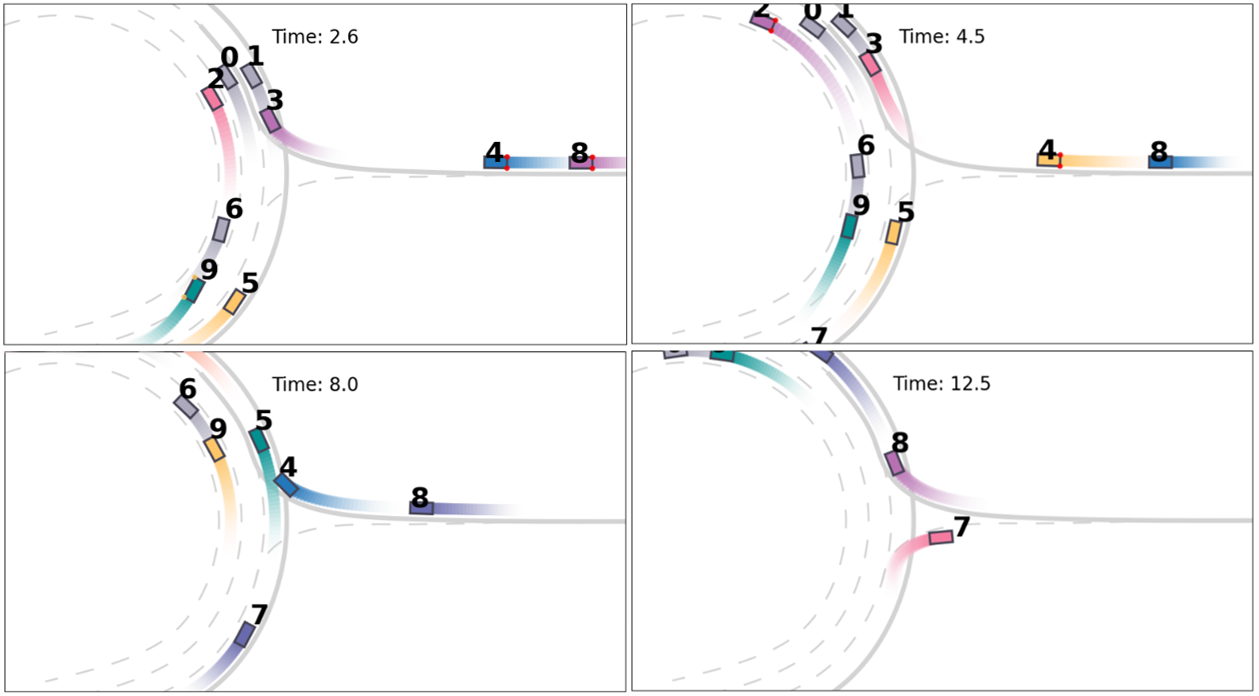}
    \caption{\rmfamily Snapshots of the Roundabout-Case 1 simulation.}
    \label{fig:test2.2_snapshot_1}
    \vspace{-10pt}
\end{figure}

\begin{figure}[t]
    \centering
    \includegraphics[width=0.98\linewidth]{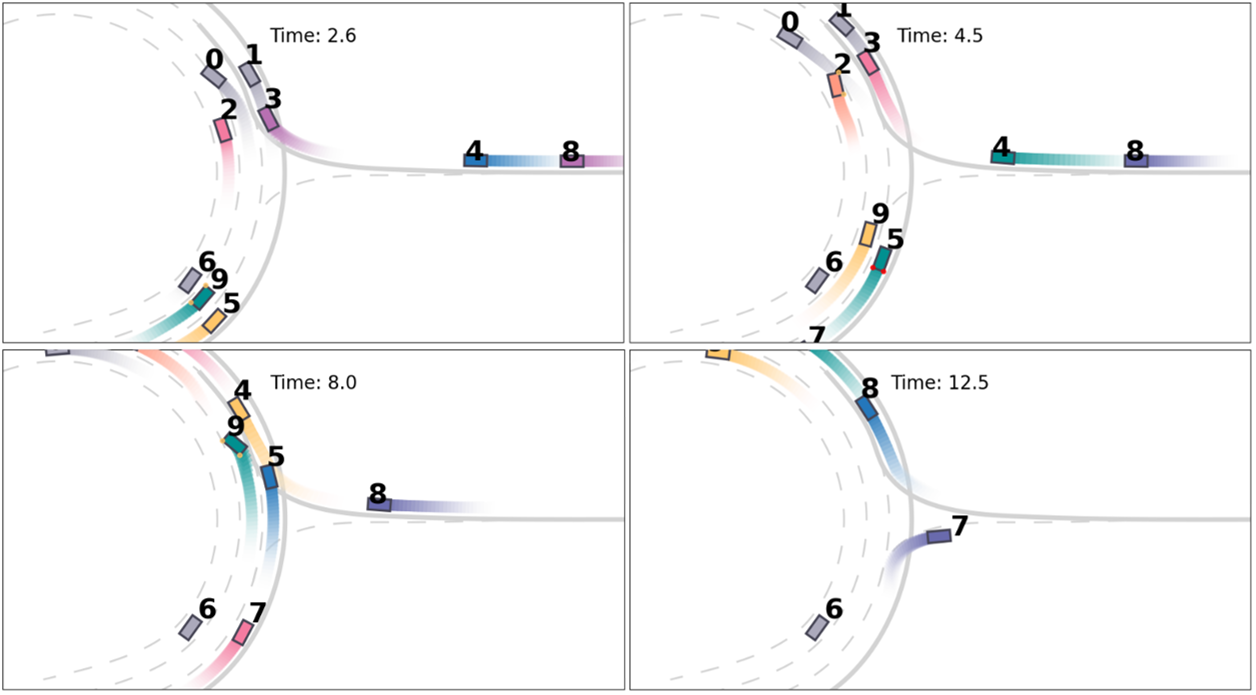}
    \caption{\rmfamily Snapshots of the Roundabout-Case 2 simulation.}
    \label{fig:test2.2_snapshot_2}
    \vspace{-10pt}
\end{figure}

\begin{figure}[tb]
    \centering
    \includegraphics[width=0.8\linewidth]{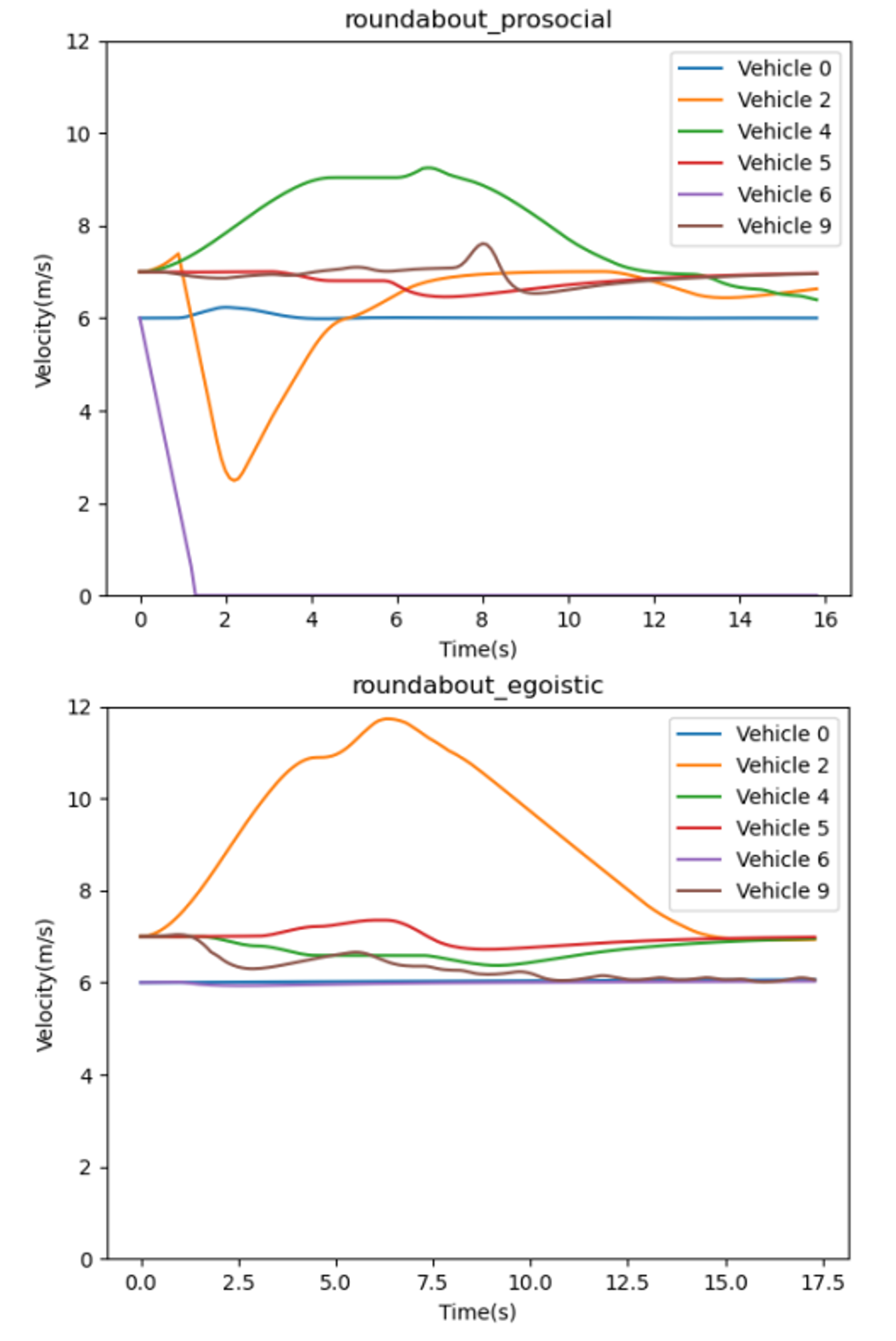}
    \caption{\rmfamily The velocity profiles of vehicle flow in the simulation process for Roundabout-Case 1 and 2.}
    \label{fig:test2.2_res}
    \vspace{-10pt}
\end{figure}

\subsection{Illustration of Traffic Flow Diversity}
We conduct several closed-loop simulations in the freeway and roundabout scenarios to illustrate how our proposed framework can bring diversity to the vehicle flow at the social levels. Meanwhile, we demonstrate the framework’s ability to complete the Overtake driving intention and the responsiveness when facing other vehicles’ unanticipated actions.

\subsubsection{Four-lane Expressway}

The first test scenario is shown in Fig. \ref{fig:test2.1_scenario}. In this freeway scenario containing a four-lane expressway, there are five AVs of $V_i~(i = 0,1,2,3,4)$ controlled by our framework. $V_0$ and $V_3$ keep driving in their current lane. $V_2$ and $V_4$ intent to change to the adjacent lane. $V_1$ aims to overtake $V_0$. In the initial state of the vehicle flow, $V_2$ and $V_3$ are driving side by side in the adjacent lanes, so $V_3$ will interfere with $V_2$’s lane-changing intention. $V_0$ and $V_1$ constitute an overtake pair. Since $V_4$ is far away from $V_0$ and $V_1$, it does not affect the driving behavior of $V_0$ and $V_1$. The output of our grouping algorithm is consistent with the above analysis, which divides the vehicle flow into three groups: the first group only has $V_4$, the second group contains $V_0$ and $V_1$, and the third group contains $V_2$ and $V_3$.

In Freeway-Case 1, all the vehicles maintain their default SVO angular preferences with $ \varphi_i =\pi/4~(i=0,1,2,3,4)$, which means they behave in a prosocial manner. In Freeway-Case 2, we change
the manners of $V_1$ and $V_2$ to be egoistic with the SVO angular preferences modified as $\varphi_i =0~(i=1,2)$, whereas the other vehicles remain unchanged. The distances to leading vehicles (space headway) and the velocity profiles of $V_0$ to $V_3$ in each case are shown in Fig. \ref{fig:test2.1_res}. $T_1$ and $T_2$ indicate the moment $V_2$ and $V_1$ complete their driving intentions in Freeway-Case 1. $T_1^{\prime}$ and $T_2^{\prime}$ indicate the moment $V_2$ and $V_1$ complete their driving intentions in Freeway-Case 2. Thus, after $T_1$ and $T_1^{\prime}$, $V_2$ becomes the leading vehicle of $V_3$ through lane-changing; after $T_2$ and $T_2^{\prime}$, $V_1$ becomes the leading vehicle of $V_0$ through overtaking. It is shown that in Freeway-Case 1, $V_1$ and $V_2$ take more time to complete their driving intentions. Even though the speeds of $V_0$ and $V_3$ are relatively slow, $V_1$ and $V_2$ still choose to change their lanes after being far away from $V_0$ and $V_3$ to ensure safety and least influence on other vehicles. In Freeway-Case 2, since $V_1$ and $V_2$ are only concerned about their rewards, the search trees provide the solutions for them to complete their driving intentions as quickly as possible. Thus, $V_3$ must slow down to make space for $V_2$ to complete its lane-changing at $T_1$, and $V_1$ does not accelerate that much as in Freeway-Case 1 to keep a large distance from $V_0$. Therefore, the velocities of $V_0$ and $V_3$ shown in Fig. \ref{fig:test2.1_res} exhibit a rapid decrease, indicating that their driving behaviors are significantly influenced by $V_1$ and $V_2$.

\begin{table*}[tb]
    \caption{\rmfamily Results of cases with different SVO settings for traffic flow diversity.}
    \label{tab:test2.1_res}
    \centering 
    \resizebox{0.8\textwidth}{!}{%
    \begin{tabular}{cccccc}
    \toprule
     Case & Min. Space Headway $(m)$ & Avg. Finish Time $(s)$ & Max. Finish Time $(s)$ & Avg. Velocity $(m/s)$	\\
    \midrule    
    Freeway-Case 1   & $13.54$ & $5.3$ & $9.3$ & $6.77$\\
    Freeway-Case 2   & $7.59$  & $4.3$ & $7.8$ & $6.68$\\
    \bottomrule
    \end{tabular}
    }
\end{table*}

The experimental indicators of the simulation process are shown in Table \ref{tab:test2.1_res}, which are consistent with the above analysis. In Freeway-Case 2, the average time taken by vehicles to complete their driving intentions is shorter. However, due to egoistic driving behaviors, the minimum distance between vehicles in the flow is smaller, which is prone to risks.

\begin{figure*}[tb]
    \centering
    \includegraphics[width=0.75\linewidth]{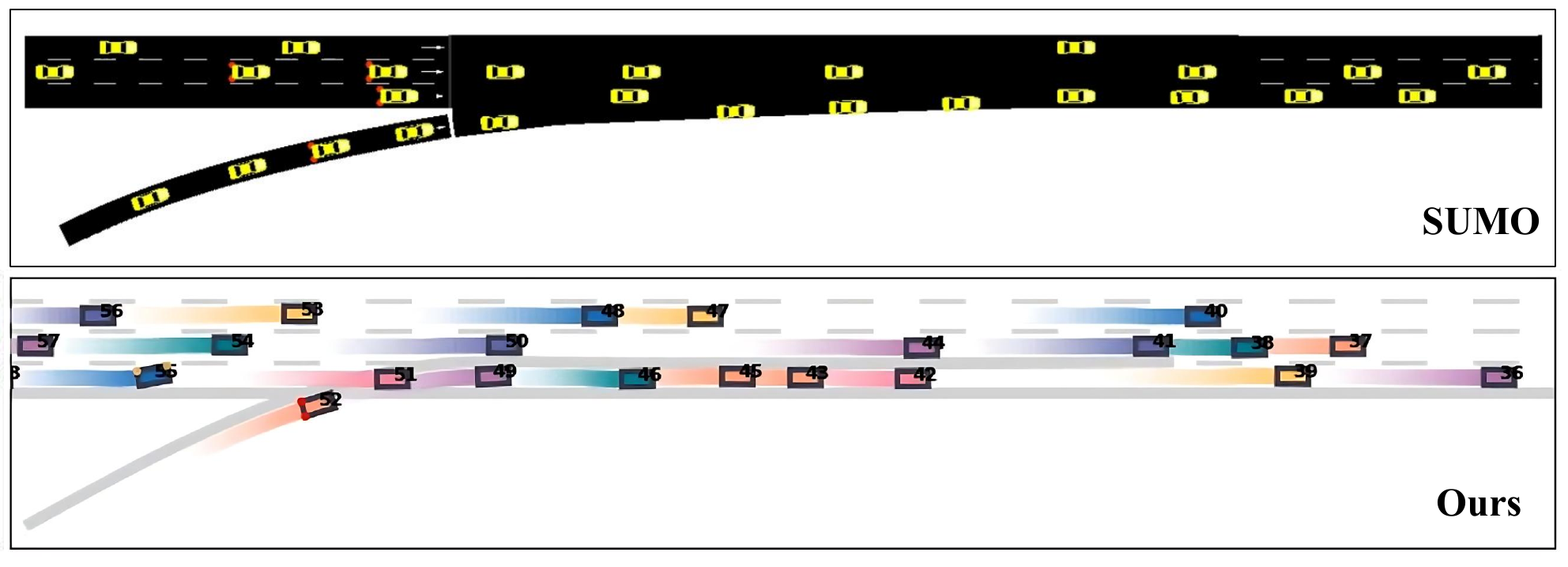}
    \caption{\rmfamily \textcolor{black}{Snapshots of the simulation process at 58.0 $s$. The input traffic volume is set as 3600 $veh/h$.}}
    \label{fig:test3.1_snapshot}
    \vspace{-10pt}
\end{figure*}

\subsubsection{Unsignalized Three-lane Roundabout Entrance}
The second test scenario is shown in Fig. \ref{fig:test2.2_scenario}. In this roundabout scenario with an unsignalized three-lane roundabout entrance, there are three uncontrolled HVs, $V_0$, $V_1$, and $V_6$. The remaining seven AVs are controlled by our framework. In the initial state of the vehicle flow, $V_3$, $V_4$, and $V_8$ are on the ramp and intend to merge into the main road. $V_2$ and $V_5$ are assigned with the driving intention of Change\_Lane\_Right and Change\_Lane\_Left respectively. $V_7$ is intent to exit the main road and enter the ramp. It should be mentioned that $V_4$ and $V_5$ almost simultaneously reach the intersection point of the main road and ramp, resulting in an interaction game between these two vehicles in the following time steps. Our grouping algorithm divides this initial flow into six groups.

In Roundabout-Case 1, all the HVs follow the predicted trajectories, which means they keep driving in their current lane. The SVO angular preference of $V_5$ is set as 0, and other vehicles maintain the default value, which means $V_5$ behaves in an egoistic manner. The traffic trajectories generated by our framework are shown in Fig. \ref{fig:test2.2_snapshot_1}, where the controlled AVs in the flow are denoted by colored trajectories and uncontrolled HVs by grey trajectories. Except for HVs, vehicles with the same color belong to the same group.
As shown in Fig. \ref{fig:test2.2_snapshot_1}, the closed-loop experimental results show that $V_2$ and $V_9$ successfully complete their driving intentions in their interactions with HVs. Fig. \ref{fig:test2.2_res} illustrates the change in the velocity of vehicles. Due to the slow speed of $V_0$ and the fact that the human driver does not slow down to make space for $V_2$, $V_2$ takes acceleration actions and completes the lane-changing after being far away from $V_0$. $V_9$ completes its lane-changing by slowing down and follows after $V_6$. As for the interaction game between $V_4$ and $V_5$, due to the egoistic driving behavior of $V_5$, it does not slow down to give way to $V_4$. Thus, $V_4$ decelerates in advance and follows behind $V_5$ after merging into the main road.

In Roundabout-Case 2, the driving behavior of $V_4$ is altered to be egoistic, and other vehicles maintain the default. Besides, we explore what happens when the uncontrollable HVs’ trajectories are inconsistent with the predicted trajectories input into the decision-making module. We modify the trajectory of $V_0$ to change to the left lane and bring $V_6$ an emergency stop, which simulates the abrupt lane-changing without a turn signal and sudden vehicle failure in the real scene respectively. Through the above case, the initial decision-making results will guide $V_2$ to accelerate and $V_9$ to turn left, creating dangerous situations.

The traffic trajectories generated by our closed-loop framework are shown in Fig. \ref{fig:test2.2_snapshot_2}. Thanks to the high-frequency replanning in our planning module, $V_2$ and $V_9$ abort their original actions from the decision module when they realize that the adjacent vehicles do not cooperate with them to avoid the potential collision. $V_2$ performs a deceleration to make space for $V_0$, and continues to complete its lane-changing in the next decision-making loop. $V_9$ keeps its original lane, and changes to the left lane after being far away from $V_6$. As for the interaction game between $V_4$ and $V_5$, despite $V_4$’s backward initial position, to obtain higher rewards, $V_4$ accelerates to enter the Merge Zone first. Unlike Roundabout-Case 1, this time $V_5$ must yield to $V_4$. Thus, $V_5$ decelerates in advance and follows behind $V_4$ after $V_4$ merging into the main road.

The comparison between Roundabout-Cases 1 and  2 emphasizes the importance of integrating the planning module into the framework, which can quickly respond to changes in the road and resolve potential collisions between vehicles. Besides, the cases above demonstrate that by adjusting the vehicles' social behaviors, our framework can bring interpretable diversity to the traffic flow and even generate extremely dangerous interaction scenarios.

\subsection{Long-term Closed-loop Simulation}
With the help of the two-stage closed-loop modules, our framework ensures a real-time response to the dynamic environment, which is of great significance for long-term multi-vehicle simulation. In this subsection, we conduct a comparative experiment with SUMO \cite{lopez2018microscopic} to display the efficient long-term decision-making ability of our framework. We also construct a road network containing multiple scenarios to demonstrate how our framework can be used to generate realistic large-scale traffic flows with complex interactions.

\begin{figure}[tb]
    \centering
    \includegraphics[width = 0.9\linewidth]{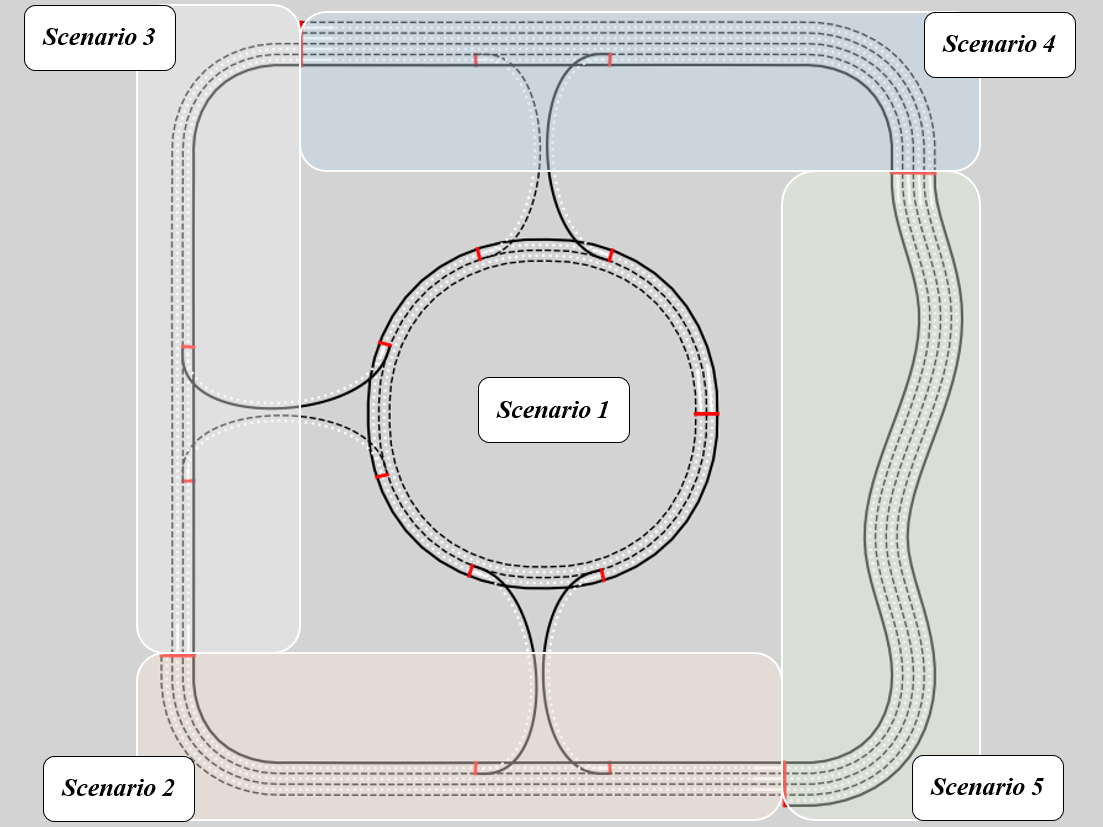}
    \caption{\rmfamily The road network containing five scenarios designed for the long-term large-scale traffic flows simulation.}
    \label{fig:test3.2_scenario}
    \vspace{-10pt}
\end{figure}

\begin{figure}[tb]
    \centering
    \includegraphics[width=1.0\linewidth]{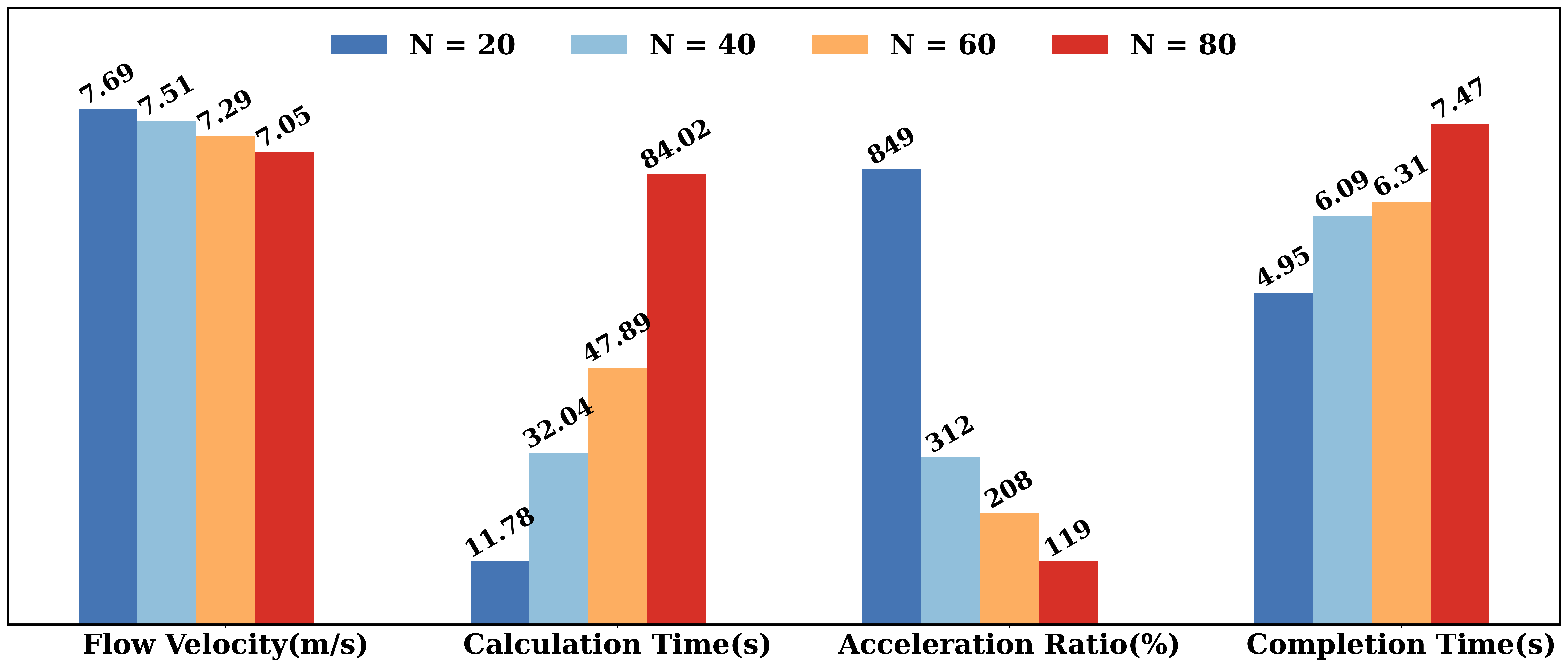}
    \caption{\rmfamily \textcolor{black}{Experimental indicators of the simulation in the road network at different numbers of vehicles.
    The indicators in the horizontal axis (from left to right) are: (1) average velocity, (2) calculation time, (3) acceleration ratio, (4) average time taken for AVs to complete their driving intentions.}}
    \label{fig:test3.2_res}
    \vspace{-5pt}
\end{figure}

\begin{figure}[tb]
    \centering
    \includegraphics[width=0.75\linewidth]{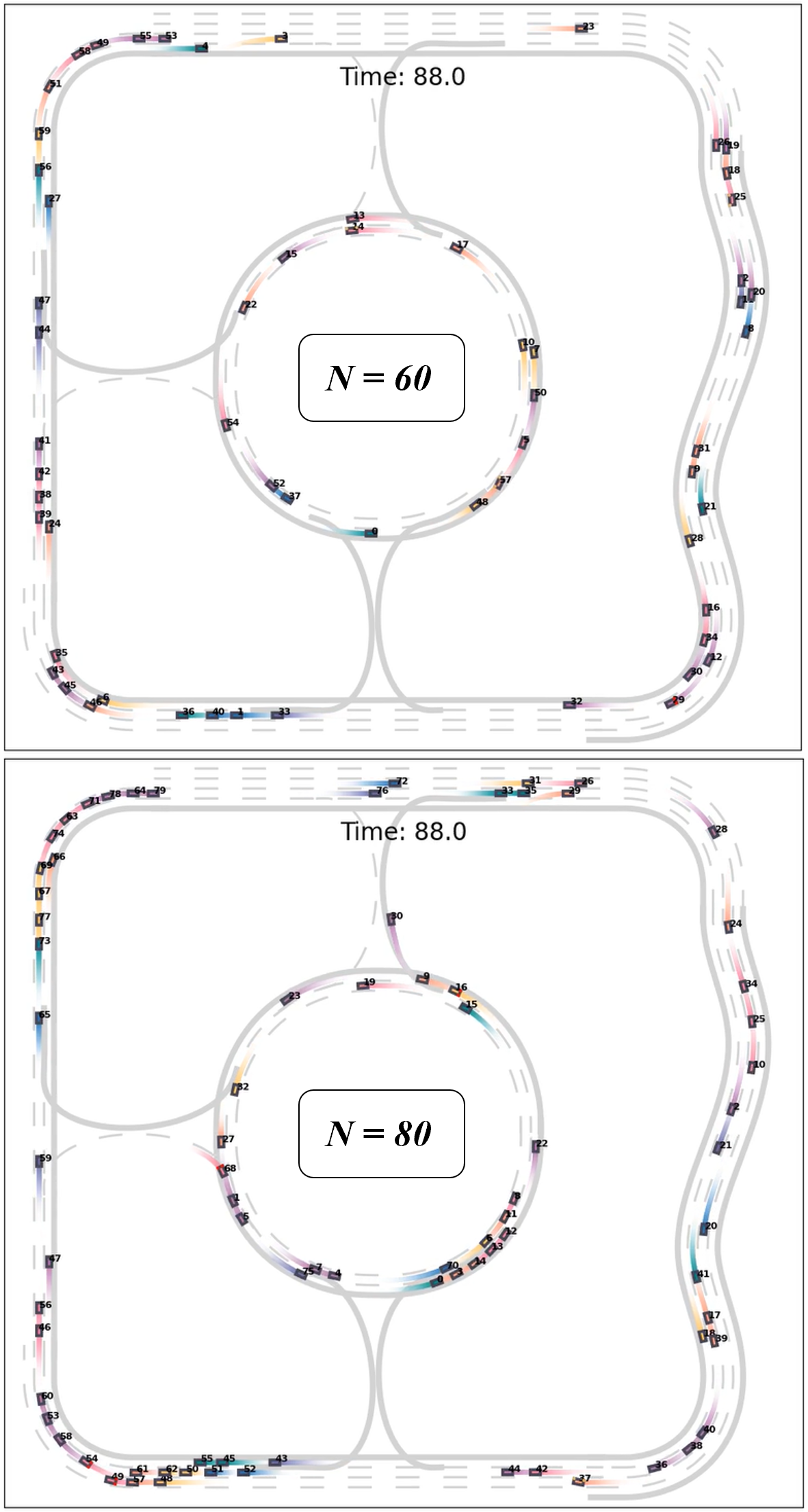}
    \caption{\rmfamily Snapshots of the simulation process at simulation time 88 $s$ with 60 and 80 vehicles.}
    \label{fig:test3.2_snapshots}
    \vspace{-10pt}
\end{figure}

\subsubsection{Comparison with SUMO}
The first test is conducted in the same ramp scenario as Section \ref{sec:test1} under different input traffic volumes. The traffic volume is set to 1800, 2400, and 3600 $veh/h$, which means that vehicles enter the scenario with time gaps of 2.0 $s$, 1.5 $s$, and 1.0 $s$, respectively. A generated vehicle will randomly depart from the beginning of one of the three main road lanes or the ramp lane with the same probability. Its initial velocity is randomly and evenly selected from $[5,7]~m/s$, and its target speed is 9 $m/s$. The total length of the main road is 200 $m$, and the vehicles' travel time is defined as the time they take from entering the scenario to arriving at the end of the main road. In addition, vehicles generated on the main road have the probability of 20\% to be assigned the intention to change to the adjacent lane, thus constituting vehicle flow with complex interactions.

\textcolor{black}{
Besides the implementation of our proposed framework, we replicate the same merging scenario in SUMO and generate vehicle routing information under different traffic volumes based on the above settings. Since our framework does not need predefined right-of-way rules, we set equal priority values for both the main road and the ramp in SUMO to enable a configuration-independent comparison. In SUMO, merging behavior is governed by conflict zones derived from network’s lane connection topology and geometry, where vehicles follow a local first-in-first-out policy. The longitudinal car-following model and lateral lane-changing model in SUMO are kept as default. The simulation is conducted using SUMO v1.23 with a total runtime of 100\,s.
}

\begin{table*}[tb]
    \caption{\rmfamily \textcolor{black}{Results of the comparative experiment with SUMO.}}
    \label{tab:test3.1_res}
    \centering
    \resizebox{0.7\textwidth}{!}{
    \begin{tabular} {c|cc|cc|cc} 
    \hline
    \multirow{2}{*}{\makecell[l]{Volume $(veh/h)$}} & \multicolumn{2}{l|}{\makecell[l]{Avg. Velocitiy $(m/s)$}} & \multicolumn{2}{l|}{\makecell[l]{Avg. Space Headway $(m)$}} & \multicolumn{2}{l}{\makecell[l]{Avg. Travel Time $(s)$}} \\ \cline{2-7}
             & Ours             & SUMO    & Ours              & SUMO      & Ours              & SUMO \\ \hline
    1800      & \textbf{8.57}    & 8.45    & \textbf{37.41}    & 36.15     & 23.08    & \textbf{20.48}  \\
    2400      & \textbf{8.57}    & 8.12    & \textbf{34.73}    & 30.66      & 22.87    & \textbf{21.18}  \\
    3600      & \textbf{8.40}    & 7.29    & \textbf{36.91}    & 27.28      & \textbf{23.54}  & 23.00  \\ \hline
    \end{tabular}
    }
    \vspace{-10pt}
\end{table*}

\textcolor{black}{
Table~\ref{tab:test3.1_res} presents the comparison results. In all settings, our method consistently achieves higher average velocity, larger space headway, and stable travel time. Under high traffic volume (3600 $(veh/h)$), our framework maintains an average speed of $8.40\,m/s$ and a headway of $36.91\,m$, while SUMO’s performance declines to $7.29\,m/s$ and $27.28\,m$, respectively. This demonstrates the robustness of our decision-making algorithm in preserving stable and efficient traffic flow under congested conditions.
}

\textcolor{black}{
Figure~\ref{fig:test3.1_snapshot} further illustrates these differences. SUMO tends to exhibit abrupt speed changes near merge zones, including sudden braking and sharp lane changes, which lead to short headways and congestion. In contrast, our framework employs group-based decision-making to model interactive behaviors and achieve smoother, more coordinated merging. It also eliminates the need for predefined lane priorities, thereby improving simulation stability and realism in complex multi-agent scenarios.
}

\subsubsection{Large-scale Road Network}
The second test is conducted in a road network containing five scenarios shown in Fig. \ref{fig:test3.2_scenario}. There is a two-lane roundabout with three entrances and exits in the center, which is labeled as Scenario 1. These entrances and exits connect to the three-lane, two-lane, and four-lane expressways, respectively, which are labeled as Scenario 2, 3, and 4. Scenario 5 is a four-lane expressway connecting Scenario 2 and 4. There are two lane-drop bottleneck sections and six on-ramp merging sections in this road network. 

Vehicles are generated in these five scenarios randomly to form the initial flow. Their initial velocities are randomly selected from $[6,8]~m/s$, and their target speeds are randomly selected from $[8,10]~m/s$. In the initial state, whenever a vehicle enters a new scenario, it will be assigned a driving intention randomly. For vehicles in the lanes of the main road of these five scenarios, their optional driving intentions include changing to the adjacent lane, overtaking the front vehicle, and entering the ramp. It should be mentioned that the vehicles facing the lane-drop bottleneck are definitely intent on changing to the right lane.

The simulation is conducted four times under different numbers of vehicles for 100 $s$. The vehicles’ average velocity and the average time to complete the driving intentions are analyzed to evaluate the performance of our framework. The results are presented in Fig. \ref{fig:test3.2_res}.
As the number of vehicles increases from 20 to 80, the road network becomes congested, leading to a decrease in average velocity. The interactions between vehicles become more and more complex, causing vehicles to take much time to complete their driving intentions. However, the average velocity remains above 7 $m/s$ in all the tests, indicating that no severe congestion is formed in the simulation process. The average time spent completing the driving intention remains within a reasonable range, which is attributed to the closed-loop architecture of our framework. Even if the decision-making fails in certain complex scenarios, subsequent re-decision and re-planning procedures can always help the vehicle efficiently complete its decision intention.
We also present the actual calculation time to generate the trajectories of the vehicle flows. The acceleration ratio in Fig. \ref{fig:test3.2_res} refers to the ratio of simulation time to the calculation time. By adopting the group-based MCTS algorithm, our framework achieves computational efficiency in large-scale traffic flow.

The snapshots of the simulation process are provided in Fig. \ref{fig:test3.2_snapshots}. All the tests generate long-term and collision-free trajectories successfully, which can be used to generate diverse traffic flows for various scenarios. Since all the vehicles' actions are interpretable and can be adjusted at both the social interaction and individual habit levels, our framework can also be utilized to perform data augmentation of the existing traffic datasets with various driving styles \cite{zang2024drivers}.

\section{Conclusion}
\label{sec:conclusion}

In this paper, we present TrafficMCTS, a novel framework for generating realistic and diverse traffic flows in a closed-loop manner. Our framework employs a novel group-based MCTS method that enhances computational efficiency by grouping vehicles according to their interaction potential. Furthermore, our framework incorporates Social Value Orientation (SVO) to model the social preferences and behaviors of different vehicles. TrafficMCTS ensures that the generated traffic flows are safe, kinematic-feasible, and adaptive to dynamic environments. 
Through experiments, we have shown that our approach surpasses existing methods in terms of computing time, planning success rate, and intention completion time.

\textcolor{black}{
In the future, we expect TrafficMCTS to serve as a traffic flow generator for interactive simulations of intelligent transportation systems. We also aim to extend our research to capture classic traffic phenomena such as traffic waves and capacity drop.
While the current framework demonstrates robustness and interpretability in multi-agent interaction modeling, it takes a fixed set of SVO parameters as input to emulate driver behavior. This static configuration may limit the model’s ability to capture contextual or situational variations in driver intent. As a future direction, we plan to incorporate adaptive behavior modeling mechanisms, such as dynamic SVO tuning or context-aware intent recognition, to further enhance the realism and behavioral diversity of generated traffic flows.
}

\bibliographystyle{IEEEtran}
\bibliography{reference}

 \balance
\end{document}